**Consistency regularization-based Deep Polynomial Chaos Neural Network Method for Reliability Analysis**

Xiaohu Zheng[a,b], Wen Yao[b,*], Yunyang Zhang[b], Xiaoya Zhang[b]

[a] College of Aerospace Science and Engineering, National University of Defense Technology, Changsha, 410073, China

[b] Defense Innovation Institute, Chinese Academy of Military Science, Beijing, 100071, China

[*] Corresponding author. E-mail: wendy0782@126.com

**ABSTRACT** Polynomial chaos expansion (PCE) is a powerful surrogate model-based reliability analysis method. Generally, a PCE model with a higher expansion order is usually required to obtain an accurate surrogate model for some complex non-linear stochastic systems. However, the high-order PCE increases the number of labeled data required for solving the expansion coefficients. To alleviate this problem, this paper proposes a consistency regularization-based deep polynomial chaos neural network (Deep PCNN) method, including the low-order adaptive PCE model (the auxiliary model) and the high-order polynomial chaos neural network (the main model). The expansion coefficients of the main model are parameterized into the learnable weights of the polynomial chaos neural network, realizing iterative learning of expansion coefficients to obtain more accurate high-order PCE models. The auxiliary model uses a proposed consistency regularization loss function to assist in training the main model. The consistency regularization-based Deep PCNN method can significantly reduce the number of labeled data in constructing a high-order PCE model without losing accuracy by using few labeled data and abundant unlabeled data. A numerical example validates the effectiveness of the consistency regularization-based Deep PCNN method, and then this method is applied to analyze the reliability of two aerospace engineering systems.

## 1. Introduction

Aerospace engineering always has uncertain factors like environmental fluctuations and manufacturing deviation [1]. These factors will affect the performance of an aerospace vehicle. Thus, reliability analysis is extremely important and essential for designing an aerospace vehicle [2,3]. Monte Carlo simulation (MCS) is an effective and simple approach for reliability analysis. However, MCS needs a large number of samples to obtain an accurate failure probability. In aerospace engineering, the cost of obtaining large amounts of aerospace vehicle performance data is very expensive and unacceptable. To reduce the number of samples, the first-order and second-order reliability methods [4] construct the local approximation of the system performance function at the most probable point (MPP). However, the MPP is hard to find exactly, which will lead to an inaccurate estimation of failure probability. Therefore, many surrogate model-based reliability analysis methods have been studied in recent years, such as polynomial chaos expansion (PCE) [5-7], deep neural network (DNN) [8,9], Kriging method [10,11], etc. Among these methods, PCE is a powerful stochastic expansion method for constructing a surrogate model.

PCE can accurately describe the randomness of random variables with any arbitrary distribution based on polynomial chaos theory. The PCE modeling includes two parts, i.e., constructing the orthogonal polynomial basis and solving the expansion coefficients. The constructing method of the orthogonal polynomial basis is very mature. Xiu et al. [5] proposed the generalized polynomial chaos (gPC) to construct the orthogonal polynomial basis of the random variable obeying the common probability distribution like Gaussian, Gamma, Beta, Uniform, etc. For the random variable obeying the arbitrary probability distribution, Oladyshkin et al. [6,7] proposed the arbitrary polynomial chaos expansion (aPC) that uses the raw moments of the random variable to construct the orthogonal polynomial basis. After constructing the orthogonal polynomial basis, the PCE model is built by calculating the sum of the product of the expansion coefficients and the orthogonal polynomial basis. Generally, the expansion coefficients can be solved by the projection method [12] or the regression method [13]. **For some complex non-linear stochastic systems, a PCE model with a higher expansion order is usually required to obtain an accurate surrogate model. However, the higher the expansion order, the more expansion coefficients need to be solved, resulting in the increase of labeled data for solving the expansion coefficients.**

Some sparse PCE (SPCE) approaches [14-16] are studied for solving the above problem. The common SPCE approaches like the orthogonal matching pursuit SPCE [17], the least angle regression SPCE [18], and the subspace pursuit SPCE [19] detect the important orthogonal polynomial basis automatically and then only solve the expansion coefficients of the important terms. Although the SPCE can effectively reduce the number of samples for solving the expansion coefficients, the precision of the surrogate model will be affected. Another good technical route is to use some unlabeled data to solve the expansion coefficients with the deep neural network (DNN), solving the cost problem of the labeled data. Two typical DNN-based PCE methods are the physics-informed neural network aPC (PINN-aPC) [20] method and the deep adaptive arbitrary polynomial chaos expansion (Deep aPCE) method [21]. Based on some physical knowledge like partial differential equations and boundary conditions, the PINN-aPC method [20] can use some unlabeled data to calculate the partial differential equation loss and the boundary condition loss. Then, the parameters of DNN are learned to solve the expansion coefficients. However, the physical knowledge of some stochastic systems is difficult or even impossible to acquire, which means the PINN-aPC method cannot be applied to this type of problem. Unlike the PINN-aPC method, the Deep aPCE method [21] calculates the adaptive aPC property loss by some unlabeled data to assist in learning the parameters of DNN. However, the Deep aPCE method only has excellent performance in solving the expansion coefficients of low-order PCE models. For a high-order PCE model, the Deep aPCE method needs a more complex DNN to solve the expansion coefficients, requiring more labeled data to train DNN. **Thus, how to use fewer data to solve the expansion coefficients of high-order PCE without losing any accuracy is a problem worth solving.**

In deep learning, consistency regularization [22,23] has been studied for reducing the number of labeled data. Consistency regularization usually uses two different DNNs (DNN-1 and DNN-2) to solve a problem based on limited labeled data and abundant unlabeled data. In the training stage, the predictions of DNN-1 on the unlabeled data serve as corresponding pseudo labels. Then, DNN-2 can take full advantage of abundant unlabeled data by forcing the consistency between the predictions of DNN-2 and pseudo labels. Therefore, this paper adopts consistency regularization to reduce the number of labeled data required for solving the expansion coefficients of high-order PCE. However, the traditional solving approaches of PCE expansion coefficients cannot directly use consistency regularization. To draw this limitation, this paper proposes a consistency regularization-based deep polynomial chaos neural network (Deep PCNN) method for reducing the number of labeled data. The Deep PCNN consists of two models, i.e., the main model and the auxiliary model. By parameterizing expansion coefficients as the learnable weights, the main model constructs a high-order PCE model as a polynomial chaos neural network. The auxiliary model is built to be a low-order adaptive PCE model by a DNN, where the DNN solves the adaptive expansion coefficients of the low-order adaptive PCE model. During the training process, the consistency regularization loss function is proposed to supervise the main model based on the pseudo labels generated by the auxiliary model on the unlabeled data. To accelerate the convergence of the parameters of the Deep PCNN, this paper proposes the expansion coefficient initialization strategy based on the properties of PCE [7]. After training the Deep PCNN, the MCS can be straightforwardly performed on the main model to analyze the reliability of the aerospace engineering systems.

The main innovations of the consistency regularization-based Deep PCNN method are presented in the following: **1) The Deep PCNN parameterizes the expansion coefficients of the high-order PCE model into the learnable weights of the polynomial chaos neural network, realizing iterative learning of expansion coefficients to obtain more accurate high-order PCE models; 2) The consistency regularization-based Deep PCNN method uses the auxiliary model to assist in training the main model by the proposed consistency regularization loss function, making the unlabeled data available for solving the expansion coefficients of higher-order PCE; 3) The consistency regularization-based Deep PCNN method is applied to construct the surrogate model of the aerospace engineering system, significantly reducing the number of labeled data without losing any accuracy.** The rest of this paper is organized as follows. In section 2, the related theories, including PCE and DNN, are introduced. The Deep PCNN and the consistency learning algorithm are proposed in section 3. Compared with the existing PCE methods, the effectiveness of the proposed Deep PCNN method is validated by a numerical example in section 4. In section 5, the proposed Deep PCNN method is applied to analyze the reliability of two aerospace engineering systems.

## 2. Related theory

*2.1. Polynomial chaos expansion*

**PCE definition.** For a engineering system, suppose that the performance function is $Y = g(X)$, where $X = [X_1, X_2, \cdots, X_d]^\mathrm{T}$ is the system input random variable (a $d$-dimensional vector). A $p$-order PCE model [24] approximates this performance function by a weighted linear combination of the multi-dimensional orthogonal basis $\{\Phi_1(\xi), \Phi_2(\xi), \cdots, \Phi_M(\xi)\}$, i.e.,

$$Y \approx y^{(p)}(\xi) = \sum_{i=1}^{M} c_i \Phi_i(\xi), \tag{1}$$

where $\xi = (\xi_1, \xi_2, \cdots, \xi_d)^\mathrm{T}$ is the corresponding variable of the input random variable $X$ in the standard random space, $c_i$ ($i = 1, 2, \cdots, M$) is the expansion coefficient, and $M = (d+p)!/d!p!$. Generally, the expansion coefficients are solved by the projection method [12] or the regression method [13]. The construction approach of the multi-dimensional orthogonal basis is introduced in the following.

**Constructing multi-dimensional orthogonal basis.** Assume that $X$ is a random univariable obeying the probability distribution $f(X)$. If $f(X)$ is a common probability distribution like Gaussian distribution, Gamma distribution, Beta distribution, etc., the correspondence of the type of orthogonal polynomial to the type of random variable is studied in the reference [5]. If $f(X)$ is not the common probability distribution, the random variable $X$ is usually transformed to its closest common distribution by the Rosenblatt or Nataf transformation. After transforming $X$ into a standard random variable $\xi$, the univariate orthogonal basis $\{\phi^{(0)}(\xi), \phi^{(1)}(\xi), \cdots, \phi^{(p)}(\xi)\}$ is constructed based on orthogonal polynomial.

In addition to the above univariate orthogonal basis constructing method, the aPC [7] use the raw moment of random univariable to construct the univariate orthogonal basis. Firstly, the random univariable $X$ is centralized and standardized by linear transformation $\xi = (X - \mu)/\sigma$, where $\mu$ and $\sigma$ are the mean and standard deviation of $X$, respectively. Then, the polynomial $\phi^{(j)}(\xi)$ of degree $j$ ($j = 0, 1, \cdots, p$) can be constructed by

$$\phi^{(j)}(\xi) = \sum_{m=0}^{j} a_m^{(j)} \xi^m, \tag{2}$$

where $a_m^{(j)}$ is the $m$th ($m = 0, 1, \cdots, j$) coefficient of $\phi^{(j)}(\xi)$. According to this reference [7], the coefficient $a_m^{(j)}$ is calculated by the raw moment $\mu_\xi^{(\beta)}$ ($\beta = 0, 1, \cdots, 2j-1$) of random univariable $\xi$, i.e.,

$$\begin{bmatrix} \mu_\xi^{(0)} & \mu_\xi^{(1)} & \cdots & \mu_\xi^{(j)} \\ \mu_\xi^{(1)} & \mu_{\xi_k}^{(2)} & \cdots & \mu_\xi^{(j+1)} \\ \vdots & \vdots & \vdots & \vdots \\ \mu_\xi^{(j-1)} & \mu_\xi^{(j)} & \cdots & \mu_\xi^{(2j-1)} \\ 0 & 0 & \cdots & 1 \end{bmatrix} \begin{bmatrix} a_0^{(j)} \\ a_1^{(j)} \\ \vdots \\ a_{j-1}^{(j)} \\ a_j^{(j)} \end{bmatrix} = \begin{bmatrix} 0 \\ 0 \\ \vdots \\ 0 \\ 1 \end{bmatrix}, \tag{3}$$

where the raw moment $\mu_\xi^{(\beta)}$ is calculated by

$$\mu_\xi^{(\beta)} = \int \xi^\beta d\Gamma(\xi). \tag{4}$$

According to the above univariate orthogonal basis constructing method, the multi-dimensional orthogonal basis $\{\Phi_1(\boldsymbol{\xi}), \Phi_2(\boldsymbol{\xi}), \cdots, \Phi_M(\boldsymbol{\xi})\}$ can be obtained by the univariate orthogonal basis $\{\phi_k^{(0)}(\xi_k), \phi_k^{(1)}(\xi_k), \cdots, \phi_k^{(p)}(\xi_k)\}$ of random variable $\xi_k$ ($k=1,2,\cdots,d$), i.e.,

$$\Phi_i(\boldsymbol{\xi}) = \prod_{k=1}^{d} \phi_k^{(s_i^k)}(\xi_k), \tag{5}$$

where $s_i^k$ is a multivariate index that contains the individual univariate basis combinatoric information [7, 21].

***Properties of PCE.*** The PCE's expansion coefficients $\{c_1, c_2, \cdots, c_M\}$ in Eq.(1) have two properties [7]:

**1) Property I.** The first expansion coefficient $c_1$ is equal to the mean $E(Y)$ of the model output $Y$, i.e.,
$$E(Y) = c_1. \tag{6}$$

**2) Property II.** The sum of squares of expansion coefficients $\{c_2, c_3, \cdots, c_M\}$ is equal to the variance of the model output $Y$, i.e.,

$$D(Y) = \sum_{i=2}^{M}(c_i)^2. \tag{7}$$

## 2.2. Deep neural network

DNN [25] includes an input layer (green circle), many hidden layers (yellow circle), and an output layer (red circle). Suppose that DNN has $L$ hidden layers. As shown in Fig. 1, $n_\alpha$ ($\alpha=1,2,\cdots,L$) denotes the number of neurons in the $\alpha$th hidden layer. For the input $\boldsymbol{x} = [x_1, x_2, \cdots, x_r]^T$, the output $\hat{\boldsymbol{y}} = [\hat{y}_1, \hat{y}_2, \cdots, \hat{y}_v]$ of DNN can be expressed as the formula

$$\hat{\boldsymbol{y}} = \mathcal{F}\left\{\mathcal{F}\left[\cdots \mathcal{F}\left[\mathcal{F}(\boldsymbol{x}^T \boldsymbol{W}_1 + \boldsymbol{b}_1)\boldsymbol{W}_2 + \boldsymbol{b}_2\right]\cdots\right]\boldsymbol{W}_L + \boldsymbol{b}_L\right\}\boldsymbol{W}_{L+1} + \boldsymbol{b}_{L+1}, \tag{8}$$

where $\mathcal{F}(\cdot)$ is the nonlinear activation function, $\boldsymbol{W} = \{\boldsymbol{W}_1, \boldsymbol{W}_2, \cdots, \boldsymbol{W}_L, \boldsymbol{W}_{L+1}\}$ and $\boldsymbol{b} = \{\boldsymbol{b}_1, \boldsymbol{b}_2, \cdots, \boldsymbol{b}_L, \boldsymbol{b}_{L+1}\}$ are the weights and the bias, respectively. In Fig. 1, $\theta_3^2$ means represents the weights and bias associated with the third neuron in the second hidden layer, and other symbols have similar meanings. Thus, $\{\theta_1^1, \theta_2^1, \cdots, \theta_{n_1}^1\} = \{\boldsymbol{W}_1, \boldsymbol{b}_1\}$, $\{\theta_1^2, \theta_2^2, \cdots, \theta_{n_2}^2\} = \{\boldsymbol{W}_2, \boldsymbol{b}_2\}$, $\cdots$, $\{\theta_1^L, \theta_2^L, \cdots, \theta_{n_L}^L\} = \{\boldsymbol{W}_L, \boldsymbol{b}_L\}$, $\{\theta_1^{L+1}, \theta_2^{L+1}, \cdots, \theta_v^{L+1}\} = \{\boldsymbol{W}_{L+1}, \boldsymbol{b}_{L+1}\}$. Denoted that the parameter $\theta$ of DNN is $\theta = \{\boldsymbol{W}, \boldsymbol{b}\}$.

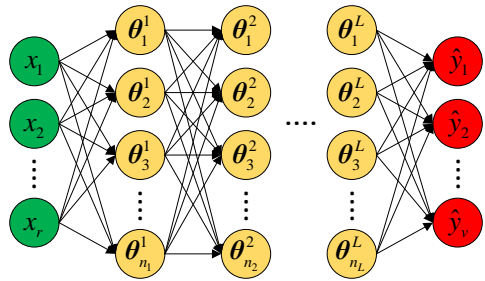

**Fig. 1.** Deep neural network.

Given $N$ labeled data $\{(\boldsymbol{x}_l, \boldsymbol{y}_l) | l = 1, 2, \cdots, N\}$, the loss function $\mathcal{L}(\boldsymbol{x}, \boldsymbol{y}; \boldsymbol{\theta})$ can be

$$\mathcal{L}(\boldsymbol{x}, \boldsymbol{y}; \boldsymbol{\theta}) = \frac{1}{N} \sum_{l=1}^{N} \left( \|\boldsymbol{y} - \hat{\boldsymbol{y}}\|_r \right)^r, \tag{9}$$

where $\|\cdot\|_r$ denotes the $r$-norm ($r=1$ or $r=2$). Thereby, the parameter $\boldsymbol{\theta}$ is learned by minimizing the loss function $\mathcal{L}(\boldsymbol{x}, \boldsymbol{y}; \boldsymbol{\theta})$, i.e.,

$$\boldsymbol{\theta}^* = \arg\min_{\boldsymbol{\theta}} \mathcal{L}(\boldsymbol{x}, \boldsymbol{y}; \boldsymbol{\theta}). \tag{10}$$

## 3. Consistency regularization-based Deep PCNN method for reliability analysis

### 3.1. Method framework

To reduce the number of labeled data, this paper proposes the consistency regularization-based Deep PCNN method (Fig. 2) for solving the higher-order PCE model's expansion coefficients. The consistency regularization-based Deep PCNN method mainly includes two parts:

**Part I: Deep PCNN modeling.** As shown in Fig. 2, the Deep PCNN consists of two models, i.e., the main model $\mathcal{M}^{(p)}(\boldsymbol{\xi}; \boldsymbol{c})$ and the auxiliary model $\mathcal{A}^{(\tilde{p})}(\boldsymbol{\xi}; \boldsymbol{\theta})$. The main model $\mathcal{M}^{(p)}(\boldsymbol{\xi}; \boldsymbol{c})$ is a $p$-order polynomial chaos neural network built based on the expansion coefficients $\boldsymbol{c} = \{c_1, c_2, \cdots, c_M\}$ and the multi-dimensional orthogonal basis $\{\Phi_1(\boldsymbol{\xi}), \Phi_2(\boldsymbol{\xi}), \cdots, \Phi_M(\boldsymbol{\xi})\}$. The polynomial chaos neural network includes an input layer $\{\xi_1, \xi_2, \cdots, \xi_M\}$, an orthogonal polynomial neural layer $\{\Phi_1(\boldsymbol{\xi}), \Phi_2(\boldsymbol{\xi}), \cdots, \Phi_M(\boldsymbol{\xi})\}$, and an output layer with only one neuron $\hat{y}^{(p)}$. The expansion coefficients $\boldsymbol{c} = \{c_1, c_2, \cdots, c_M\}$ are parameterized to be the learnable weights of the neuron $\hat{y}^{(p)}$, realizing iterative learning of expansion coefficients to obtain more accurate high-order PCE models. Based on the multi-dimensional orthogonal basis $\{\tilde{\Phi}_1(\boldsymbol{\xi}), \tilde{\Phi}_2(\boldsymbol{\xi}), \cdots, \tilde{\Phi}_{\tilde{M}}(\boldsymbol{\xi})\}$, the auxiliary model $\mathcal{A}^{(\tilde{p})}(\boldsymbol{\xi}; \boldsymbol{\theta})$ is constructed to be a $\tilde{p}$-order adaptive PCE model by a DNN, where the DNN solves the adaptive expansion coefficients $\{\tilde{c}_1(\boldsymbol{\xi}; \boldsymbol{\theta}), \tilde{c}_2(\boldsymbol{\xi}; \boldsymbol{\theta}), \cdots, \tilde{c}_{\tilde{M}}(\boldsymbol{\xi}; \boldsymbol{\theta})\}$. Thus, the parameters to be learned in the Deep PCNN include the parameterized expansion coefficients $\boldsymbol{c}$ and the parameters $\boldsymbol{\theta}$ of DNN, denoted as $(\boldsymbol{c}, \boldsymbol{\theta})$.

**Part II: Deep PCNN training.** The consistency regularization-based Deep PCNN method adopts a small amount of labeled data $\mathcal{D}_{gd} = \{(\boldsymbol{\xi}_l^{gd}, y_l^{gd}) | l = 1, 2, \cdots, N_{gd}\}$ and abundant unlabeled data $\mathcal{D}_{ce} = \{\boldsymbol{\xi}_{l'}^{ce} | l' = 1, 2, \cdots, N_{ce}\}$ to learn the parameters $(\boldsymbol{c}, \boldsymbol{\theta})$. Two supervised loss functions $\mathcal{L}_s^{\mathcal{M}}(\boldsymbol{c})$ and $\mathcal{L}_s^{\mathcal{A}}(\boldsymbol{\theta})$ are calculated using the labeled data $\mathcal{D}_{gd} = \{(\boldsymbol{\xi}_l^{gd}, y_l^{gd}) | l = 1, 2, \cdots, N_{gd}\}$. For a unlabeled data $\xi_{l'}^{ce}$, the prediction $\hat{y}_{l'}^{ce} = \mathcal{M}^{(p)}\left(\xi_{l'}^{ce}; c\right)$ should be consistent with the prediction $\tilde{y}_{l'}^{ce} = \mathcal{A}^{(\tilde{p})}\left(\xi_{l'}^{ce}; \theta\right)$. Based on this idea, the consistency regularization loss function $\mathcal{L}_c(c)$ is proposed by calculating the mean one-norm of the predicted errors between the main model $\mathcal{M}^{(p)}(\xi; c)$ and auxiliary model $\mathcal{A}^{(\tilde{p})}(\xi; \theta)$. Referring to the properties of the adaptive aPC [21], two adaptive PCE property loss functions $\mathcal{L}_{ce}^E(\theta)$ and $\mathcal{L}_{ce}^D(\theta)$ are computed to training the DNN in the auxiliary model $\mathcal{A}^{(\tilde{p})}(\xi; \theta)$ based on abundant unlabeled data. Based on the above loss functions, the parameters $(c, \theta)$ of the Deep PCNN is learned by a small amount of labeled data and abundant unlabeled data.

During the training process, the auxiliary model $\mathcal{A}^{(\tilde{p})}(\xi; \theta)$ can learn the primary features of the stochastic system by the loss function $\mathcal{L}_{\mathcal{A}}(\theta)$. Then the auxiliary model uses the consistency regularization loss function $\mathcal{L}_c(c)$ to assist the main model in learning more detailed features of the stochastic system. In summary, the construction of a surrogate model using the Deep PCNN method includes three aspects: 1) Deep PCNN modeling, 2) constructing the loss function $\mathcal{L}(c, \theta)$, and 3) training the Deep PCNN based on the labeled data and the unlabeled data. The above three aspects are introduced in detail in sections 3.2, 3.3, and 3.4, respectively.

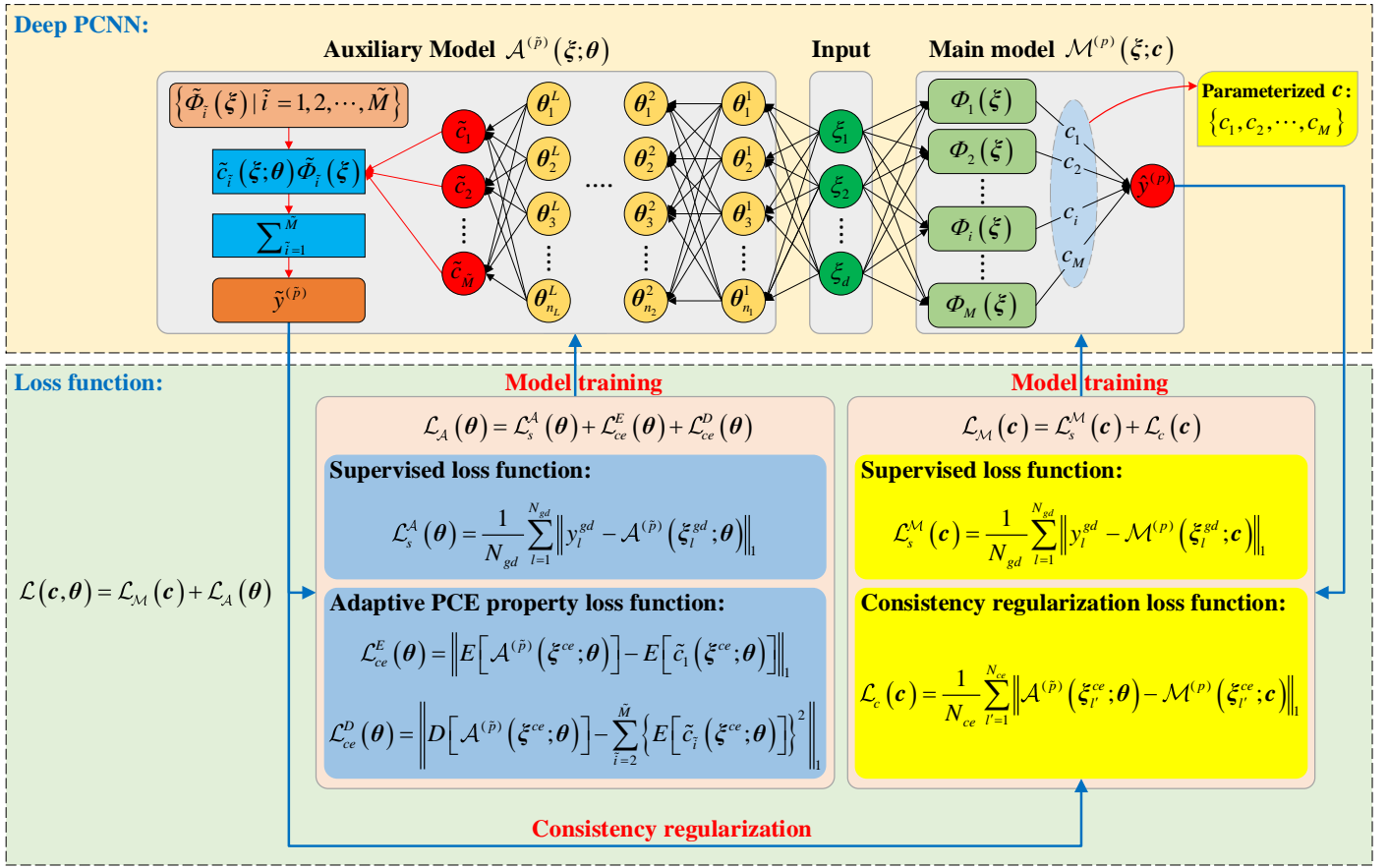

**Fig. 2.** Framework of the consistency regularization-based Deep PCNN method.

*3.2. Deep polynomial chaos neural network*

As shown in Fig. 2, the main model $\mathcal{M}^{(p)}(\boldsymbol{\xi};\boldsymbol{c})$ and the auxiliary model $\mathcal{A}^{(\tilde{p})}(\boldsymbol{\xi};\boldsymbol{\theta})$ make up the Deep PCNN. The above two models are constructed as follows.

**1) Main model** $\mathcal{M}^{(p)}(\boldsymbol{\xi};\boldsymbol{c})$

As shown in Fig. 3(a), supposed that a DNN with one hidden layer is used to approximate the performance function $Y = G(\boldsymbol{\xi})$. Supposed the parameter $\boldsymbol{\theta}_i$ of the $i$ th neuron in the hidden layer is $\{W_1^h, W_2^h, \cdots, W_d^h, b_h\}$, and the parameter $\boldsymbol{\theta}_{\hat{y}}$ of $\{W_1^{\hat{y}}, W_2^{\hat{y}}, \cdots, W_M^{\hat{y}}, b_{\hat{y}}\}$. According to Eq.(8), the calculation of the output $\hat{y}$ is

$$\begin{aligned} H_i &= \mathcal{F}\left(\sum_{q=1}^{d} W_q^h \xi_q + b_h\right), \\ \boldsymbol{H} &= (H_1, H_2, \cdots, H_i, \cdots, H_M)^{\mathrm{T}}, \\ \hat{y} &= \boldsymbol{\xi}^{\mathrm{T}} \boldsymbol{H} + b_{\hat{y}} = \sum_{i=1}^{M} W_i^{\hat{y}} H_i + b_{\hat{y}}. \end{aligned} \tag{11}$$

Inspired by the DNN output's calculation method in Eq.(11), this paper constructs the main model $\mathcal{M}^{(p)}(\boldsymbol{\xi};\boldsymbol{c})$ to be a $p$-order polynomial chaos neural network as shown in Fig. 3(b), which is modeled based on the $p$-order PCE model in Eq.(1). Compared with the DNN in Fig. 3(a), the $p$-order polynomial chaos neural network uses an orthogonal polynomial neural layer $\{\Phi_1(\boldsymbol{\xi}), \Phi_2(\boldsymbol{\xi}), \cdots, \Phi_M(\boldsymbol{\xi})\}$ without nonlinear activation function to replace the hidden layer. It is noteworthy that the orthogonal polynomial neural layer does not include any parameter to learn. Thus, the values of neurons in the orthogonal polynomial neural layer are calculated by

$$\begin{aligned} H_i &= \Phi_i(\boldsymbol{\xi}), \quad (i = 1, 2, \cdots, M), \\ \boldsymbol{H} &= (H_1, H_2, \cdots, H_i, \cdots, H_M)^{\mathrm{T}}. \end{aligned} \tag{12}$$

In the $p$-order polynomial chaos neural network, the expansion coefficients $\boldsymbol{c} = \{c_1, c_2, \cdots, c_M\}$ are parameterized to be the learnable weights of the output layer neuron as shown in Fig. 3(b). Referring to Eq.(11), the output $\hat{y}^{(p)}$ is calculated based on Eq.(12), i.e.,

$$\hat{y}^{(p)} = \boldsymbol{\xi}^{\mathrm{T}} \boldsymbol{H} = \sum_{i=1}^{M} c_i H_i = \sum_{i=1}^{M} c_i \Phi_i(\boldsymbol{\xi}). \tag{13}$$

Thus, the parameters of the main model $\mathcal{M}^{(p)}(\boldsymbol{\xi};\boldsymbol{c})$ are the parameterized expansion coefficients $\boldsymbol{c} = \{c_1, c_2, \cdots, c_M\}$.

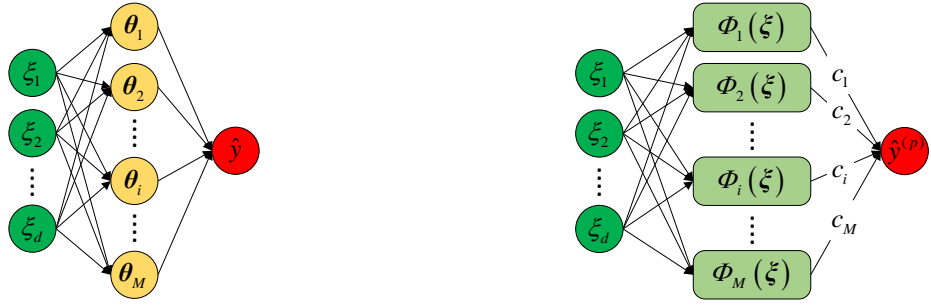

(a) The DNN with one hidden layer  (b) The $p$-order polynomial chaos neural network

**Fig. 3** The comparison between the DNN and the $p$-order polynomial chaos neural network.

**2) Auxiliary model** $\mathcal{A}^{(\tilde{p})}(\xi;\theta)$

Based on two properties of the adaptive aPC, Yao and Zheng et al. [21] proposed the Deep aPCE method that has excellent performance in solving the expansion coefficients of low-order PCE models with only a small amount of labeled data. Thus, this paper uses a $\tilde{p}$-order adaptive PCE model to build the auxiliary model $\mathcal{A}^{(\tilde{p})}(\xi;\theta)$, i.e.,

$$\tilde{y}^{(\tilde{p})} = \sum_{\tilde{i}=1}^{\tilde{M}} \tilde{c}_{\tilde{i}}(\xi;\theta)\tilde{\Phi}_{\tilde{i}}(\xi). \tag{14}$$

As shown in Fig. 2, the auxiliary model $\mathcal{A}^{(\tilde{p})}(\xi;\theta)$ uses a DNN with $L$ hidden layers to solve the adaptive expansion coefficients $\{\tilde{c}_1(\xi;\theta),\tilde{c}_2(\xi;\theta),\cdots,\tilde{c}_{\tilde{M}}(\xi;\theta)\}$. Therefore, the parameters of the auxiliary model $\mathcal{A}^{(\tilde{p})}(\xi;\theta)$ are the parameters $\theta = \{W, b\}$ of DNN.

Finally, the main model $\mathcal{M}^{(p)}(\xi;c)$ and the auxiliary model $\mathcal{A}^{(\tilde{p})}(\xi;\theta)$ form the Deep PCNN, where the expansion order $p$ is larger than $\tilde{p}$, i.e., $p > \tilde{p}$. In training the Deep PCNN, the low-order auxiliary model learns the primary features of the stochastic system to assist the high-order main model in further learning more detailed features of the stochastic system, making the main model $\mathcal{M}^{(p)}(\xi;c)$ more accurate.

*3.3. Loss function for training Deep PCNN*

To train the Deep PCNN, this paper prepares two kinds of data, i.e.,

- The labeled data $\mathcal{D}_{gd} = \{(\xi_l^{gd}, y_l^{gd}) | l = 1, 2, \cdots, N_{gd}\}$;
- The unlabeled data $\mathcal{D}_{ce} = \{\xi_{l'}^{ce} | l' = 1, 2, \cdots, N_{ce}\}$.

For the consistency regularization-based Deep PCNN method, the auxiliary model $\mathcal{A}^{(\tilde{p})}(\xi;\theta)$ is used to assist in training the main model $\mathcal{M}^{(p)}(\xi;c)$ based on the proposed loss function $\mathcal{L}(c,\theta)$, i.e.,

$$\mathcal{L}(c,\theta) = \mathcal{L}_{\mathcal{M}}(c) + \mathcal{L}_{\mathcal{A}}(\theta), \tag{15}$$

The calculations of two loss functions $\mathcal{L}_{\mathcal{M}}(c)$ and $\mathcal{L}_{\mathcal{A}}(\theta)$ are presented as follows.

**1) Loss function $\mathcal{L}_{\mathcal{M}}(c)$ for training the main model $\mathcal{M}^{(p)}(\xi;c)$**

For the input $\xi_l^{gd}$ ($l=1,2,\cdots,N_{gd}$), the corresponding prediction $\hat{y}_l^{gd}$ is predicted by the main model, i.e. $\hat{y}_l^{gd} = \mathcal{M}^{(p)}(\xi_l^{gd};c)$. Thus, the supervised loss function $\mathcal{L}_s^{\mathcal{M}}(c)$ is calculated by the mean one-norm of the errors between the ground truth values $\{y_l^{gd} \mid l=1,2,\cdots,N_{gd}\}$ and the predictions $\{\hat{y}_l^{gd} \mid l=1,2,\cdots,N_{gd}\}$, i.e.,

$$\mathcal{L}_s^{\mathcal{M}}(c) = \frac{1}{N_{gd}} \sum_{l=1}^{N_{gd}} \left\| y_l^{gd} - \hat{y}_l^{gd} \right\|_1 = \frac{1}{N_{gd}} \sum_{l=1}^{N_{gd}} \left\| y_l^{gd} - \mathcal{M}^{(p)}(\xi_l^{gd};c) \right\|_1. \tag{16}$$

Besides, the corresponding predictions $\{\hat{y}_{l'}^{ce} \mid l'=1,2,\cdots,N_{ce}\}$ of the unlabeled inputs $\{\xi_{l'}^{ce} \mid l'=1,2,\cdots,N_{ce}\}$ are obtained by the main model, where $\hat{y}_{l'}^{ce} = \mathcal{M}^{(p)}(\xi_{l'}^{ce};c)$. In the consistency regularization-based Deep PCNN method, the value $\tilde{y}_{l'}^{ce}$ predicted by the auxiliary model is regarded as the pseudo label for the input $\xi_{l'}^{ce}$. Therefore, the consistency regularization loss function $\mathcal{L}_c(c)$ is calculated by the mean one-norm of the errors between the pseudo labels $\{\tilde{y}_{l'}^{ce} \mid l'=1,2,\cdots,N_{ce}\}$ and the prediction $\{\hat{y}_{l'}^{ce} \mid l'=1,2,\cdots,N_{ce}\}$, i.e.,

$$\mathcal{L}_c(c) = \frac{1}{N_{ce}} \sum_{l'=1}^{N_{ce}} \left\| \tilde{y}_{l'}^{ce} - \hat{y}_{l'}^{ce} \right\|_1 = \frac{1}{N_{ce}} \sum_{l'=1}^{N_{ce}} \left\| \mathcal{A}^{(\tilde{p})}(\xi_{l'}^{ce};\theta) - \mathcal{M}^{(p)}(\xi_{l'}^{ce};c) \right\|_1. \tag{17}$$

In summary, the loss function $\mathcal{L}_{\mathcal{M}}(c)$ is calculated based on Eq.(16) and Eq.(17), i.e.,

$$\mathcal{L}_{\mathcal{M}}(c) = \mathcal{L}_s^{\mathcal{M}}(c) + \mathcal{L}_c(c). \tag{18}$$

In the consistency regularization-based Deep PCNN method, the loss function $\mathcal{L}_{\mathcal{M}}(c)$ is used to learn the parameterized expansion coefficients $c = \{c_1, c_2, \cdots, c_M\}$ of the main model $\mathcal{M}^{(p)}(\xi;c)$.

**2) Loss function $\mathcal{L}_{\mathcal{A}}(\theta)$ for training the auxiliary model $\mathcal{A}^{(\tilde{p})}(\xi;\theta)$**

The corresponding prediction $\tilde{y}_l^{gd}$ of the input $\xi_l^{gd}$ ($l=1,2,\cdots,N_{gd}$) is estimated by the auxiliary model, i.e. $\tilde{y}_l^{gd} = \mathcal{A}^{(\tilde{p})}(\xi_l^{gd};\theta)$. Thus, the supervised loss function $\mathcal{L}_s^{\mathcal{A}}(\theta)$ is calculated by the mean one-norm of the errors between the ground truth values $\{y_l^{gd} \mid l=1,2,\cdots,N_{gd}\}$ and the predictions $\{\tilde{y}_l^{gd} \mid l=1,2,\cdots,N_{gd}\}$, i.e.,

$$\mathcal{L}_s^{\mathcal{A}}(\boldsymbol{\theta}) = \frac{1}{N_{gd}} \sum_{l=1}^{N_{gd}} \left\| y_l^{gd} - \tilde{y}_l^{gd} \right\|_1$$
$$= \frac{1}{N_{gd}} \sum_{l=1}^{N_{gd}} \left\| y_l^{gd} - \mathcal{A}^{(\tilde{p})}\left(\boldsymbol{\xi}_l^{gd}; \boldsymbol{\theta}\right) \right\|_1. \quad (19)$$

For $N_{ce}$ unlabeled inputs $\{\boldsymbol{\xi}_{l'}^{ce} \mid l' = 1, 2, \cdots, N_{ce}\}$, the corresponding predictions $\{\tilde{y}_{l'}^{ce} \mid l' = 1, 2, \cdots, N_{ce}\}$ and the expansion coefficients $\{\tilde{\boldsymbol{c}}_{l'}(\boldsymbol{\xi}^{ce}; \boldsymbol{\theta}) \mid l' = 1, 2, \cdots, N_{ce}\}$ are computed by auxiliary model, where $\tilde{y}_{l'}^{ce} = \mathcal{A}^{(\tilde{p})}(\boldsymbol{\xi}_{l'}^{ce}; \boldsymbol{\theta})$ and $\tilde{\boldsymbol{c}}_{l'}(\boldsymbol{\xi}^{ce}; \boldsymbol{\theta}) = \{\tilde{c}_{\tilde{i}}^{l'}(\boldsymbol{\xi}; \boldsymbol{\theta}) \mid \tilde{i} = 1, 2, \cdots, \tilde{M}\}$. Thus, the mean $E[\mathcal{A}^{(\tilde{p})}(\boldsymbol{\xi}^{ce}; \boldsymbol{\theta})]$ and the variance $D[\mathcal{A}^{(\tilde{p})}(\boldsymbol{\xi}^{ce}; \boldsymbol{\theta})]$ of the predictions $\{\tilde{y}_{l'}^{ce} \mid l' = 1, 2, \cdots, N_{ce}\}$ are

$$E\left[\mathcal{A}^{(\tilde{p})}\left(\boldsymbol{\xi}^{ce}; \boldsymbol{\theta}\right)\right] = \frac{1}{N_{ce}} \sum_{l'=1}^{N_{ce}} \tilde{y}_{l'}^{ce},$$
$$D\left[\mathcal{A}^{(\tilde{p})}\left(\boldsymbol{\xi}^{ce}; \boldsymbol{\theta}\right)\right] = \frac{1}{N_{ce}} \sum_{l'=1}^{N_{ce}} \left(\tilde{y}_{l'}^{ce}\right)^2 - \left\{E\left[\mathcal{A}^{(\tilde{p})}\left(\boldsymbol{\xi}^{ce}; \boldsymbol{\theta}\right)\right]\right\}^2. \quad (20)$$

Besides, the mean $E[\tilde{c}_{\tilde{i}}(\boldsymbol{\xi}; \boldsymbol{\theta})]$ of the $\tilde{i}$ th ($\tilde{i} = 1, 2, \cdots, \tilde{M}$) expansion coefficient $\tilde{c}_{\tilde{i}}(\boldsymbol{\xi}; \boldsymbol{\theta})$ is

$$E\left[\tilde{c}_{\tilde{i}}\left(\boldsymbol{\xi}^{ce}; \boldsymbol{\theta}\right)\right] = \frac{1}{N_{ce}} \sum_{l'=1}^{N_{ce}} \tilde{c}_{\tilde{i}}^{l'}\left(\boldsymbol{\xi}^{ce}; \boldsymbol{\theta}\right). \quad (21)$$

According to two properties of adaptive expansion coefficients in the reference [21], two adaptive PCE property loss functions $\mathcal{L}_{ce}^E(\boldsymbol{\theta})$ and $\mathcal{L}_{ce}^D(\boldsymbol{\theta})$ are

$$\mathcal{L}_{ce}^E(\boldsymbol{\theta}) = \left\| E\left[\mathcal{A}^{(\tilde{p})}\left(\boldsymbol{\xi}^{ce}; \boldsymbol{\theta}\right)\right] - E\left[\tilde{c}_1\left(\boldsymbol{\xi}^{ce}; \boldsymbol{\theta}\right)\right] \right\|_1,$$
$$\mathcal{L}_{ce}^D(\boldsymbol{\theta}) = \left\| D\left[\mathcal{A}^{(\tilde{p})}\left(\boldsymbol{\xi}^{ce}; \boldsymbol{\theta}\right)\right] - \sum_{\tilde{i}=2}^{\tilde{M}} \left\{E\left[\tilde{c}_{\tilde{i}}\left(\boldsymbol{\xi}^{ce}; \boldsymbol{\theta}\right)\right]\right\}^2 \right\|_1. \quad (22)$$

In summary, the loss function $\mathcal{L}_{\mathcal{A}}(\boldsymbol{\theta})$ is calculated based on Eq.(19) and Eq.(22), i.e.,

$$\mathcal{L}_{\mathcal{A}}(\boldsymbol{\theta}) = \mathcal{L}_s^{\mathcal{A}}(\boldsymbol{\theta}) + \mathcal{L}_{ce}^E(\boldsymbol{\theta}) + \mathcal{L}_{ce}^D(\boldsymbol{\theta}). \quad (23)$$

In the consistency regularization-based Deep PCNN method, the loss function $\mathcal{L}_{\mathcal{A}}(\boldsymbol{\theta})$ is used to learn the parameters $\boldsymbol{\theta} = \{\boldsymbol{W}, \boldsymbol{b}\}$ of the auxiliary model $\tilde{y}^{(\tilde{p})} = \mathcal{A}^{(\tilde{p})}(\boldsymbol{\xi}; \boldsymbol{\theta})$.

*3.4. Consistency learning algorithm*

This section proposes a consistency learning algorithm to learning the parameters $(\boldsymbol{c}, \boldsymbol{\theta})$ of the Deep PCNN based on the proposed loss function $\mathcal{L}(\boldsymbol{c}, \boldsymbol{\theta})$. Before training the Deep PCNN, the parameters $(\boldsymbol{c}, \boldsymbol{\theta})$ need to be initialized firstly. To accelerate the convergence of the main model $\mathcal{M}^{(p)}(\xi;c)$, this section proposes the initialization strategy for the parameter $c = \{c_1, c_2, \cdots, c_M\}$ in the following.

**1) Strategy for initializing the parameter $c_1$**

According to Eq.(6), the first expansion coefficient $c_1$ is always a constant for the PCE model (Eq.(1)) with different orders ($p = 2, 3, 4, \cdots$). Besides, the low-order PCE model requires less labeled data to calculate the expansion coefficients than the high-order PCE model. Thus, a $p_{low}$-order PCE model is constructed based on the labeled data $\mathcal{D}_{gd} = \{(\xi_l^{gd}, y_l^{gd}) \,|\, l = 1, 2, \cdots, N_{gd}\}$. Supposed that the first expansion coefficient of the $p_{low}$-order PCE model is denoted as $c_1^{low}$. Then, the parameter $c_1$ is initialized by the expansion coefficient $c_1^{low}$, i.e.,

$$c_1 = c_1^{low}. \tag{24}$$

**2) Strategy for initializing the parameters $\{c_2, c_3, \cdots, c_M\}$**

For the labeled data $\mathcal{D}_{gd} = \{(\xi_l^{gd}, y_l^{gd}) \,|\, l = 1, 2, \cdots, N_{gd}\}$, the variance $D(y^{gd})$ is

$$D(y^{gd}) = \frac{1}{N_{gd}} \sum_{l=1}^{N_{gd}} (y_l^{gd})^2 - \left(\frac{1}{N_{gd}} \sum_{l=1}^{N_{gd}} y_l^{gd}\right)^2. \tag{25}$$

For $i = 2, 3, \cdots, M$, the following inequation can be obtained by Eq.(7), i.e.,

$$(c_i)^2 \leq D(y^{gd}) \quad \Rightarrow \quad -\sqrt{D(y^{gd})} \leq c_i \leq \sqrt{D(y^{gd})}. \tag{26}$$

Thus, the parameters $\{c_2, c_3, \cdots, c_M\}$ are initialized by randomly sampling from the uniform distribution with the lower boundary $-\sqrt{D(y^{gd})}$ and the upper boundary $\sqrt{D(y^{gd})}$, i.e.,

$$c_i \sim U\left(-\sqrt{D(y^{gd})}, \sqrt{D(y^{gd})}\right). \tag{27}$$

By the above initialization strategies, the parameter $c$ of the main model $\mathcal{M}^{(p)}(\xi;c)$ can get the initial values. During the training process, the labeled data $\mathcal{D}_{gd}$ and the unlabeled data $\mathcal{D}_{ce}$ are input to the Deep PCNN to get the corresponding predictions, based on which the loss function $\mathcal{L}(c, \theta)$ is calculated by Eq.(15). Then, the gradients $\partial \mathcal{L}_s^M(c)/\partial c$, $\partial \mathcal{L}_c(c)/\partial c$, $\partial \mathcal{L}_c(c)/\partial \theta$, $\partial \mathcal{L}_s^A(\theta)/\partial \theta$, $\partial \mathcal{L}_{ce}^E(\theta)/\partial \theta$, and $\partial \mathcal{L}_{ce}^D(\theta)/\partial \theta$ are derived by the chain rule for differentiating compositions of functions using automatic differentiation [26]. In the consistency learning algorithm, the gradient propagation in the Deep PCNN is shown in Fig. 4. In particular, the loss function $\mathcal{L}_\mathcal{M}(c)$ includes the parameters $c$ and $\theta$ according to Eq.(17). Thus, the gradients $\partial \mathcal{L}_c(c)/\partial c$ and $\partial \mathcal{L}_c(c)/\partial \theta$ will be obtained in the automatic differentiation process. Since the auxiliary model $\mathcal{A}^{(\tilde{p})}(\xi;\theta)$ is used to assist the training of the main model $\mathcal{M}^{(p)}(\xi;c)$, the gradient propagation of $\partial \mathcal{L}_c(c)/\partial \theta$ in the auxiliary model $\mathcal{A}^{(\tilde{p})}(\xi;\theta)$ is blocked as shown in Fig. 4. Finally, the gradients $\partial \mathcal{L}_s^A(\theta)/\partial \theta$, $\partial \mathcal{L}_{ce}^E(\theta)/\partial \theta$, and $\partial \mathcal{L}_{ce}^D(\theta)/\partial \theta$ are used to update the parameter $\theta$ of the auxiliary model $\mathcal{A}^{(\tilde{p})}(\xi;\theta)$, and the gradients $\partial \mathcal{L}_s^M(c)/\partial c$, $\partial \mathcal{L}_c(c)/\partial c$ are used to update the parameter $c$ of the main model $\mathcal{M}^{(p)}(\xi;c)$.

In this paper, $ep_{\max}$ is assumed to be the maximum training epoch. For the $ep$ th ($ep = 1, 2, \cdots, ep_{\max}$) epoch, the Adam algorithm [27] is adopted to update the parameters $(c, \theta)$ of the Deep PCNN based on five calculated gradients $\partial \mathcal{L}_s^M(c)/\partial c$, $\partial \mathcal{L}_c(c)/\partial c$, $\partial \mathcal{L}_s^A(\theta)/\partial \theta$, $\partial \mathcal{L}_{ce}^E(\theta)/\partial \theta$, and $\partial \mathcal{L}_{ce}^D(\theta)/\partial \theta$. In summary, the pseudo code of the consistency learning algorithm is shown in the following **Algorithm**.

---

**Algorithm:** Consistency learning algorithm

**Input:**
    (1) Maximum training epoch $ep_{\max}$;
    (2) Labeled data $\mathcal{D}_{gd} = \{(\xi_l^{gd}, y_l^{gd}) | l = 1, 2, \cdots, N_{gd}\}$;
    (3) Unlabeled data $\mathcal{D}_{ce} = \{\xi_{l'}^{ce} | l' = 1, 2, \cdots, N_{ce}\}$.

**Output:**
    Trained main model $\mathcal{M}^{(p)}(\xi;c)$.

1    Using the proposed strategy to initialize the parameter $c$ of the main model $\mathcal{M}^{(p)}(\xi;c)$;
2    **for** $ep = 1: ep_{\max}$ **do**
3      **for** $l = 1: N_{gd}$ **do**
4        Using the main model to calculate the predicted value $\hat{y}_l^{gd} = \mathcal{M}^{(p)}(\xi_l^{gd};c)$;
5        Using the auxiliary model to calculate the predicted value $\tilde{y}_l^{gd} = \mathcal{A}^{(\tilde{p})}(\xi_l^{gd};\theta)$;
6      **end**
7      **for** $l' = 1: N_{ce}$ **do**
8        Using the main model to calculate the predicted value $\hat{y}_{l'}^{ce} = \mathcal{M}^{(p)}(\xi_{l'}^{ce};c)$;
9        Using DNN of the auxiliary model to calculate the expansion coefficients $\tilde{c}_{l'}(\xi^{ce};\theta) = \{\tilde{c}_{\tilde{i}}^{l'}(\xi;\theta) | \tilde{i} = 1, 2, \cdots, \tilde{M}\}$;
10        Using the auxiliary model to calculate the predicted value $\tilde{y}_{l'}^{ce} = \mathcal{A}^{(\tilde{p})}(\xi_{l'}^{ce};\theta)$
11      **end**
12     Calculating the mean $E\left[\mathcal{A}^{(\tilde{p})}(\xi^{ce};\theta)\right] = 1/N_{ce} \sum_{l'=1}^{N_{ce}} \tilde{y}_{l'}^{ce}$;
13     Calculating the variance $D\left[\mathcal{A}^{(\tilde{p})}(\xi^{ce};\theta)\right] = 1/N_{ce} \sum_{l'=1}^{N_{ce}} (\tilde{y}_{l'}^{ce})^2 - \left\{E\left[\mathcal{A}^{(\tilde{p})}(\xi^{ce};\theta)\right]\right\}^2$;
14     **for** $\tilde{i} = 1: \tilde{M}$ **do**
15       Calculating the mean $E\left[\tilde{c}_{\tilde{i}}(\xi^{ce};\theta)\right] = 1/N_{ce} \sum_{l'=1}^{N_{ce}} \tilde{c}_{\tilde{i}}^{l'}(\xi^{ce};\theta)$
16     **end**
17     Calculating the loss function $\mathcal{L}(c, \theta) = \mathcal{L}_\mathcal{M}(c) + \mathcal{L}_\mathcal{A}(\theta)$;
18     Calculating the gradients $\partial \mathcal{L}_s^M(c)/\partial c$, $\partial \mathcal{L}_c(c)/\partial c$, $\partial \mathcal{L}_s^A(\theta)/\partial \theta$, $\partial \mathcal{L}_{ce}^E(\theta)/\partial \theta$, and $\partial \mathcal{L}_{ce}^D(\theta)/\partial \theta$;
19     Using the Adam algorithm to update the parameters $(c, \theta)$ of the Deep PCNN;
20    **end**

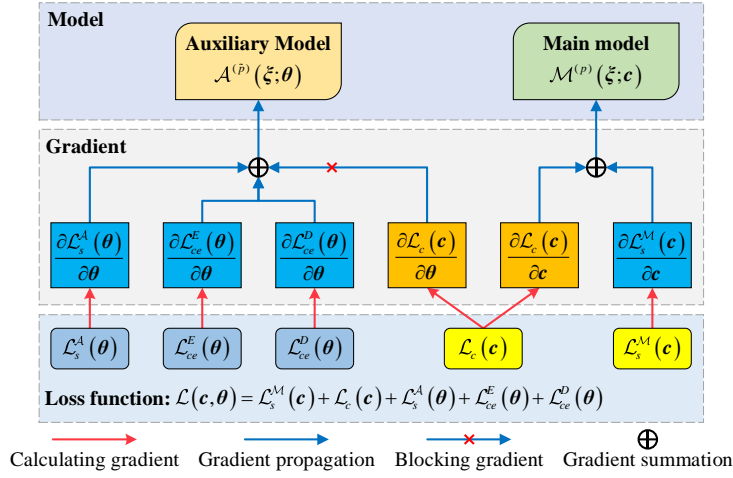

**Fig. 4** Schematic diagram of gradient propagation in the Deep PCNN.

*3.5. Deep PCNN-based reliability analysis*

According to section 3.4, the Deep PCNN can be trained by the proposed consistency learning algorithm. In particular, the trained main model $\mathcal{M}^{(p)}(\boldsymbol{\xi};\boldsymbol{c})$ is the final predictive model. Based on the testing data $\{(\boldsymbol{\xi}_{l_t}^t, y_{l_t}^t)|l_t = 1, 2, \cdots, N_t\}$, the accuracy of the trained main model $\mathcal{M}^{(p)}(\boldsymbol{\xi};\boldsymbol{c})$ is evaluated by four error indicators, i.e.,

1) the root mean square error $RMSE$

$$RMSE = \sqrt{\frac{1}{N_t} \sum_{l_t=1}^{N_t} \left[ y_{l_t}^t - \mathcal{M}^{(p)}(\boldsymbol{\xi}_{l_t}^t;\boldsymbol{c}) \right]^2}, \tag{28}$$

2) the mean absolute error $MAE$

$$MAE = \frac{1}{N_t} \sum_{l_t=1}^{N_t} \left| y_{l_t}^t - \mathcal{M}^{(p)}(\boldsymbol{\xi}_{l_t}^t;\boldsymbol{c}) \right|, \tag{29}$$

3) the mean relative error $MRE$

$$MRE = \frac{1}{N_t} \sum_{l_t=1}^{N_t} \left| \frac{y_{l_t}^t - \mathcal{M}^{(p)}(\boldsymbol{\xi}_{l_t}^t;\boldsymbol{c})}{y_{l_t}^t} \right|, \tag{30}$$

4) the R square $R^2$

$$R^2 = 1 - \frac{\sum_{l_t=1}^{N_t} \left[ y_{l_t}^t - \mathcal{M}^{(p)}(\boldsymbol{\xi}_{l_t}^t;\boldsymbol{c}) \right]^2}{\sum_{l_t=1}^{N_t} \left[ y_{l_t}^t - \bar{y}^t \right]^2}, \tag{31}$$

where the mean $\bar{y}^t$ is calculated by

$$\overline{y}^t = \frac{1}{N_t}\sum_{l_t=1}^{N_t}\mathcal{M}^{(p)}\left(\xi_{l_t}^t;\boldsymbol{c}\right). \tag{32}$$

In this paper, the failure probability $P_f$ is calculated by performing MCS on the trained main model $\mathcal{M}^{(p)}(\xi;\boldsymbol{c})$, i.e.,

$$P_f \approx \frac{1}{N_{\text{MCS}}}\sum_{l_{\text{MCS}}=1}^{N_{\text{MCS}}} IF\left[\mathcal{M}^{(p)}\left(\xi_{l_{\text{MCS}}},\boldsymbol{c}\right)\right], \tag{33}$$

where $N_{\text{MCS}}$ is the number of samples, and $IF(\cdot)$ is the indicator function. If $\mathcal{M}^{(p)}(\xi;\boldsymbol{c})<0$, $IF\left[\mathcal{M}^{(p)}\left(\xi_{l_{\text{MCS}}},\boldsymbol{c}\right)\right]=1$, otherwise $IF\left[\mathcal{M}^{(p)}\left(\xi_{l_{\text{MCS}}},\boldsymbol{c}\right)\right]=0$.

## 4. Numerical example

### 4.1. Example background

As shown in **Fig. 5**, this paper chooses the cantilever tube [28,29] to validate the effectiveness of the proposed consistency regularization-based Deep PCNN method.

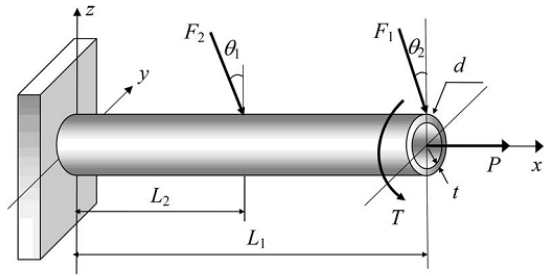

**Fig. 5** Cantilever tube [28].

The performance function $g(\boldsymbol{x},\boldsymbol{z})$ is defined to be

$$g(\boldsymbol{x},\boldsymbol{z}) = S_y - \sqrt{\sigma_x^2 + 3\tau_{zx}^2}, \tag{34}$$

where $S_y$ is the yield strength, the normal stress $\sigma_x$ is

$$\sigma_x = \frac{P + F_1\sin(\theta_1) + F_2\sin(\theta_2)}{A} + \frac{Md}{2I}, \tag{35}$$

and the torsional stress $\tau_{zx}$ is

$$\tau_{zx} = \frac{Td}{4I}. \tag{36}$$

In Eq. (35),

$$A = \frac{\pi}{4}[d^2 - (d-2t)^2], \tag{37}$$

$$I = \frac{\pi}{64}[d^4 - (d-2t)^4], \tag{38}$$

and the bending moment $M$ is calculated by

$$M = F_1 L_1 \cos(\theta_1) + F_2 L_2 \cos(\theta_2) \tag{39}$$

The random variables are shown in Table 1. In this paper, the parameters $\theta_1$ and $\theta_2$ are set to be 5 deg and 10 deg, respectively. If $g(\boldsymbol{x},\boldsymbol{z}) \leq 0$, the cantilever tube fails.

**Table 1** Nine random variables for the cantilever tube numerical example.

| Random variable | Mean (Lower boundary) | Standard deviation (Upper boundary) | Distribution |
| --- | --- | --- | --- |
| $t$ | 5 mm | 0.1 mm | Normal |
| $d$ | 42 mm | 0.5 mm | Normal |
| $L_1$ | 119.75 mm | 120.25 mm | Uniform |
| $L_2$ | 59.75 mm | 60.25 mm | Uniform |
| $F_1$ | 3.0 kN | 0.3 kN | Normal |
| $F_2$ | 3.0 kN | 0.3 kN | Normal |
| $P$ | 12.0 kN | 1.2 kN | Gumbel |
| $T$ | 90.0 Nm | 9.0 Nm | Normal |
| $S_y$ | 220 MPa | 22.0 MPa | Normal |

*4.2. Constructing PCNN model*

In this numerical example, a 2-order adaptive PCE model is used to construct the auxiliary model, and the adaptive expansion coefficients are solved by a DNN with 55 outputs and five hidden layers, where the neuron numbers of five hidden layers are 32, 64, 128, 64, and 64, respectively. Besides, the activation function is the $\text{ReLU}(x)$. The main model is a 4-order, 5-order, or 6-order polynomial chaos neural network. For three polynomial chaos neural networks, the numbers of neurons in the orthogonal polynomial neural layer are 715, 2002, and 5005, respectively. The dataset $\mathcal{D}_{ce}$ consists of $2 \times 10^5$ unlabeled data, and the number $N_{gd}$ of labeled data is 90, 200, and 400, respectively. The maximum training epoch $ep_{\max}$ is set to be 20000. In this numerical example, the results by the MCS method ($1 \times 10^6$ runs) are assumed to be the real results.

*4.3. Comparison between Deep PCNN and Deep aPCE*

Based on 400 labeled data and $2 \times 10^5$ unlabeled data, this section compares the accuracies of two surrogate models constructed by the Deep aPCE and the proposed Deep PCNN, respectively. The Deep aPCE is built by a 2-order aPC model. For the proposed Deep PCNN, the main model is established using a 6-order aPC model, and the previous Deep aPCE is used as the auxiliary model. For two surrogate models, the relative errors of standard deviation, skewness and kurtosis, and three error indicators (RMSE, MAE,

MRE) are shown in Fig. 6. The relative errors of the Deep aPCE on the standard deviation, the skewness, and the kurtosis are much larger than the proposed Deep PCNN. In particular, the relative error of the proposed Deep PCNN on the skewness is only 1.7140%, while the Deep aPCE arrives at 11.1014%. For the errors of two surrogate models, the RMSE, MAE and MRE of the Deep aPCE are greater than the proposed Deep PCNN. Besides, the R square $R^2$ of the proposed Deep PCNN is 0.99999892, which is less than the R square $R^2$ (0.9999817) of the Deep aPCE.

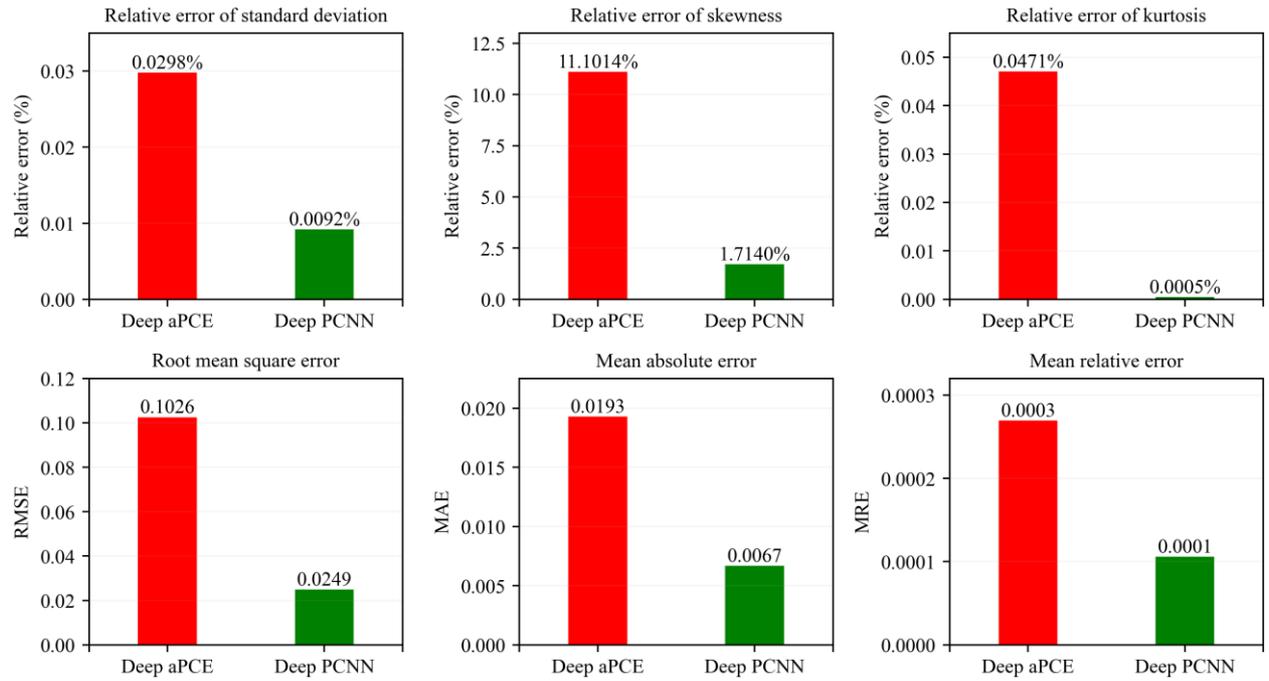

Fig. 6 The relative errors of standard deviation, skewness and kurtosis, and three error indicators (RMSE, MAE, MRE). The Deep aPCE is built by a 2-order aPC model, and the main model of the Deep PCNN is established using a 6-order aPC model.

According to the above analysis, two conclusions can be summarized as follows: 1) Based on the same amount of labeled data, the proposed Deep PCNN can construct a more accurate surrogate model than the Deep aPCE; 2) By the consistency regularization, the low-order auxiliary model assists the high-order main model to learn more detailed features of the stochastic system, sequentially constructing a more accurate surrogate model.

*4.4. Comparison between Deep PCNN and existing PCE methods*

Based on the UQLab-V2.0-104 [30], the cantilever tube example is also solved by the Ordinary Least-Squares PCE (OLS-PCE) method, the Orthogonal Matching Pursuit Sparse PCE (OMP-SPCE) method [17], and the Least Angle Regression Sparse PCE (LAR-SPCE) method [18]. The first four statistical moments and their relative errors are shown in Table 2, and the bold results indicate the best results.

For $p=4$, although four methods can correctly calculate the mean of the performance function value, the relative errors of three existing PCE methods on the standard deviation, the skewness, and the kurtosis are larger than the proposed Deep PCNN.

Among three existing PCE methods, the relative error on the standard deviation of the OLS-PCE ( $N_{gd} = 390$ ) is minimal (0.01587%). However, it is still larger than the proposed Deep PCNN (0.01178%). Besides, the minimal relative errors on the skewness and the kurtosis is the OMP-SPCE ( $N_{gd} = 590$ ), i.e., 8.2132% on the skewness and 0.004423% on the kurtosis, while the relative error on the skewness and the kurtosis of the proposed Deep PCNN ( $N_{gd} = 90$ ) are only 2.7507% and 0.003104%, respectively. For $p = 5, 6$, the relative errors on the standard deviation, the skewness, and the kurtosis of the proposed Deep PCNN are still minimum in all methods. It is noteworthy that the proposed Deep PCNN always uses the fewest labeled data.

Table 2 The first four statistical moments and their relative errors in the cantilever tube numerical example.

| $p$ | Method | $N_{gd}$ | Mean | S.D. | Skewness (R.E.) | Kurtosis (R.E.) |
|---|---|---|---|---|---|---|
| - | MCS | $1 \times 10^6$ | 85.78 | 23.9463 | -0.005410 | 2.9976 |
| 4 | OLS-PCE | 390 | 85.78 | 23.9425 (0.01587%) | -0.004108 (24.0619%) | 2.9962 (0.049763%) |
|   | OMP-SPCE | 590 | 85.78 | 23.9526 (0.02630%) | -0.004965 (8.2132%) | 2.9975 (0.004423%) |
|   | LAR-SPCE | 690 | 85.78 | 23.9519 (0.02356%) | -0.001065 (80.3155%) | 2.9971 (0.018136%) |
|   | **Deep PCNN** | **90** | 85.78 | **23.9434 (0.01178%)** | **-0.005260 (2.7507%)** | **2.9976 (0.003104%)** |
| 5 | OLS-PCE | 900 | 85.78 | 23.9510 (0.01983%) | -0.005884 (8.7757%) | 2.9978 (0.006856%) |
|   | OMP-SPCE | 2500 | 85.78 | 23.9418 (0.01857%) | -0.004891 (9.5791%) | 2.9975 (0.004170%) |
|   | LAR-SPCE | 3400 | 85.78 | 23.9344 (0.04944%) | -0.001081 (80.0203%) | 2.9971 (0.018618%) |
|   | **Deep PCNN** | **200** | 85.78 | **23.9429 (0.01421%)** | **-0.005208 (3.7065%)** | **2.9976 (0.000527%)** |
| 6 | OLS-PCE | 2060 | 85.78 | 23.9500 (0.01546%) | -0.005061 (6.4368%) | 2.9985 (0.029554%) |
|   | OMP-SPCE | 3600 | 85.78 | 23.9751 (0.02090%) | -0.004882 (9.7556%) | 2.9975 (0.004362%) |
|   | LAR-SPCE | 6000 | 85.78 | 23.9406 (0.02349%) | -0.001091 (79.8380%) | 2.9971 (0.018373%) |
|   | **Deep PCNN** | **400** | 85.78 | **23.9441 (0.00921%)** | **-0.005316 (1.7140%)** | **2.9976 (0.000493%)** |

S.D. = Standard deviation      R.E. = Relative error

To further validate the accuracy of the proposed Deep PCNN, four error indicators, the failure probability $P_f$ and its relative estimation error rate $\varepsilon_{P_f}$ are also calculated as shown in Table 3. For $p = 4$, the minimum *RMSE*, *MAE*, and *MRE* in three existing PCE methods are 0.1588, 0.0756, and $1.0159 \times 10^{-3}$, respectively. However, the *RMSE*, *MAE*, and *MRE* of the proposed Deep PCNN are only 0.0438, 0.0287, and $0.4045 \times 10^{-3}$, respectively. Apparently, the *RMSE*, *MAE*, and *MRE* of the proposed Deep PCNN are smaller than three existing PCE methods. The maximum $R^2$ of three existing PCE methods is 0.99995602, while the proposed Deep PCNN is 0.99999665. Besides, the estimated failure probability $P_f$ of the proposed Deep PCNN is $1.840 \times 10^{-4}$, which is the closest value to the MCS result ( $1.850 \times 10^{-4}$ ). Compared with the failure probability estimated by MCS, the relative estimation error rates of the failure probability $P_f$ for four methods are 5.4054% (LAR-SPCE), 4.8649% (OLS-PCE), 2.7027% (OMP-SPCE), 0.5405% (Deep PCNN), respectively. Thus, the proposed Deep PCNN can provide the most accurate estimation of the failure probability $P_f$ in all methods. For $p = 5, 6$, the *RMSE*, *MAE*, and *MRE* of the proposed Deep PCNN are still minimum in all methods, and the proposed Deep PCNN still has the largest $R^2$. In particular, the 6-order Deep PCNN can estimate the failure probability $P_f$ without any relative estimation error.

Table 3 Four error indicators and the estimated failure probability in the cantilever tube numerical example.

| $p$ | Method | $N_{gd}$ | RMSE | MAE | MRE ($\times 10^{-3}$) | $R^2$ | $P_f$ ($\times 10^{-4}$) | $\varepsilon_{P_f}$ |
|---|---|---|---|---|---|---|---|---|
| - | MCS | $1\times 10^6$ | - | - | - | - | 1.850 | - |
| 4 | OLS-PCE | 390 | 0.1676 | 0.0756 | 1.0159 | 0.99995103 | 1.760 | 4.8649% |
|   | OMP-SPCE | 590 | 0.1588 | 0.1248 | 1.5977 | 0.99995602 | 1.800 | 2.7027% |
|   | LAR-SPCE | 690 | 0.3729 | 0.2782 | 3.6541 | 0.99975751 | 1.750 | 5.4054% |
|   | **Deep PCNN** | **90** | **0.0438** | **0.0287** | **0.4045** | **0.99999665** | **1.840** | **0.5405%** |
| 5 | OLS-PCE | 900 | 0.3234 | 0.0817 | 1.1331 | 0.99981755 | 1.800 | 2.7027% |
|   | OMP-SPCE | 2500 | 0.1584 | 0.1242 | 1.6020 | 0.99995623 | 1.790 | 3.2432% |
|   | LAR-SPCE | 3400 | 0.3696 | 0.2769 | 3.6325 | 0.99976172 | 1.740 | 5.9459% |
|   | **Deep PCNN** | **200** | **0.0139** | **0.0089** | **0.1382** | **0.99999966** | **1.840** | **0.5405%** |
| 6 | OLS-PCE | 2060 | 0.2788 | 0.0498 | 0.6819 | 0.99986445 | 1.810 | 2.1622% |
|   | OMP-SPCE | 3600 | 0.1581 | 0.1242 | 1.5907 | 0.99995639 | 1.800 | 2.7027% |
|   | LAR-SPCE | 6000 | 0.3696 | 0.2780 | 3.6419 | 0.99976182 | 1.740 | 5.9459% |
|   | **Deep PCNN** | **400** | **0.0249** | **0.0067** | **0.1057** | **0.99999892** | **1.850** | **0.0000%** |

The absolute error boxplots of four methods are shown in Fig. 7. According to Fig. 7, the medians and the interquartile ranges (IQRs) of the LAR-SPCE and the OMP-SPCE are always larger than the OLS-PCE and the Deep PCNN. It means that the overall accuracies of the latter two methods are higher than the former two methods. Compared with the OLS-PCE and the Deep PCNN, the former two methods perform the orthogonal basis sparse operation. Thus, the orthogonal basis sparse operation of the LAR-SPCE and the OMP-SPCE affect the accuracy of the surrogate model. Besides, the medians and the IQRs of the OLS-PCE and the Deep PCNN decrease gradually as the order increases, but the medians and the IQRs of the proposed Deep PCNN are always far less than the OLS-PCE. What's more, the proposed Deep PCNN always adopts less labeled data than the OLS-PCE. Especially, the OLS-PCE ($N_{gd}=2060$) uses more than five times as much labeled data as the proposed Deep PCNN ($N_{gd}=400$) for $p=6$. Therefore, the proposed Deep PCNN can construct a more accurate surrogate model than the OLS-PCE with less labeled data.

By the result analyses for Table 2, Table 3, and Fig. 7, the proposed Deep PCNN can use less labeled data to build a more accurate surrogate model without any complex orthogonal basis sparse operation than the existing PCE methods. Thus, the proposed Deep PCNN can solve the existing PCE methods' shortcoming, i.e., needing to increase the expansion order to improve the accuracy of the surrogate model but causing more labeled data to solve the expansion coefficients.

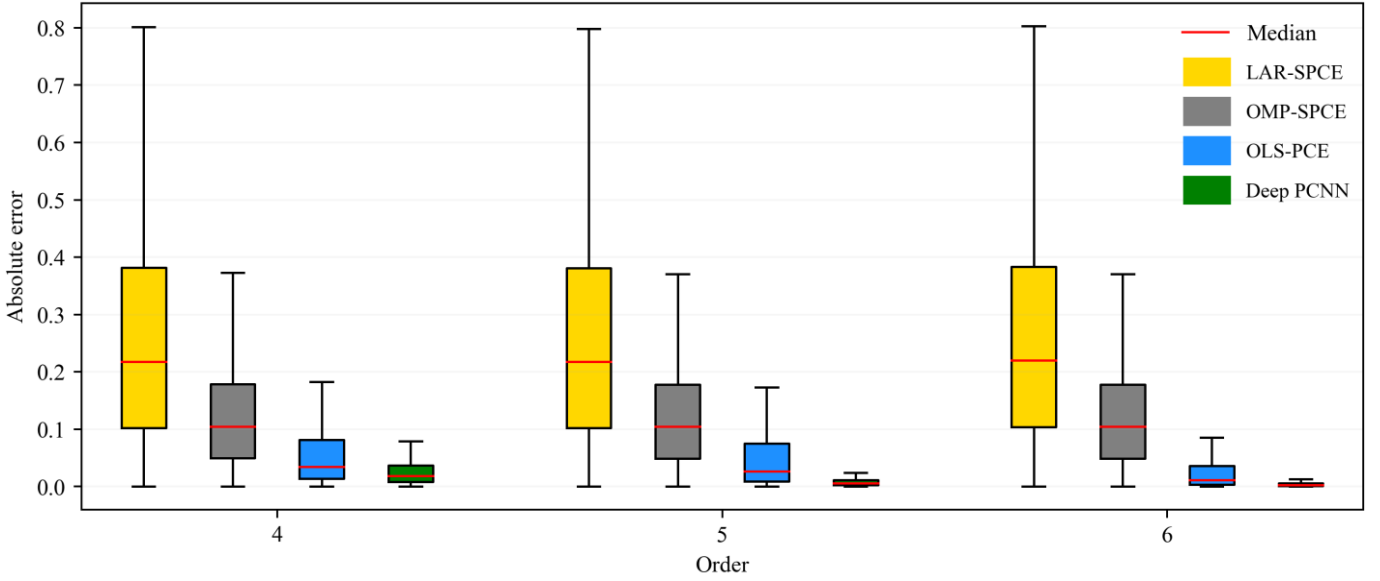

Fig. 7 The absolute error boxplots of four methods under different orders in the cantilever tube numerical example.

## 5. Engineering Applications

This section uses two engineering cases to validate the effectiveness of the proposed consistency regularization-based Deep PCNN method. Two applications use the 2-order ($\tilde{p}=2$) adaptive PCE model to construct the auxiliary model $\mathcal{A}^{(\tilde{p})}(\boldsymbol{\xi};\boldsymbol{\theta})$. Besides, the values of random input variables are sampled by the Latin Hypercube Sampling. The relevant codes by Python are available on this website[1]. The OLS-PCE method, the OMP-SPCE method [17], the LAR-SPCE method [18], and the Subspace Pursuit Sparse PCE (SP-SPCE) [19] are achieved by the UQLab-V2.0-104 [30].

*5.1. Application 1: Reliability analysis for microsatellite conceptual design*

*5.1.1. Engineering background of application 1*

"Tiantuo-1" satellite is a microsatellite that integrates basic functional components on a single circuit board, as shown in Fig. 8. Combined with the development experience of the "Tiantuo-1" satellite, the microsatellite generally uses many new technologies and commercial components to reduce its volume and mass, which greatly increases the uncertainty of the system. Besides, the design and production cycle of microsatellites is short, and the ground test procedure is simplified. Thus, it is essential to analyze the system reliability in the microsatellite conceptual design stage.

---

[1] https://github.com/Xiaohu-Zheng/Deep-Polynomial-Chaos-Neural-Network-Method

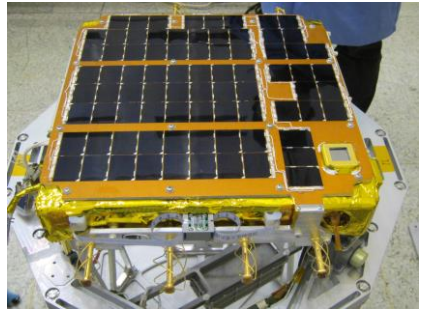

**Fig. 8** "Tiantuo-1" satellite

Since the mass is directly and positively related to satellite cost, the microsatellite mass is one of the design indexes to consider in the conceptual design stage. Generally, the microsatellite mass is influenced by the following seven parameters [31], i.e., the sun-synchronous circle orbit height $h_o$, the charge-coupled device (CCD) camera focal length $f_c$, the cross profile width (vertical flight direction) $b_{sat}$, the satellite height $l_{sat}$, the satellite wall thickness $t_{sat}$. The *Application 1* studies a conceptual microsatellite with the following design scheme, i.e., $h_o = 600 \text{ km}$, $f_c = 280 \text{ mm}$, $b_{sat} = 800 \text{ mm}$, $l_{sat} = 700 \text{ mm}$, and $t_{sat} = 5 \text{ mm}$. For this conceptual microsatellite, if the microsatellite mass is more than 183 kg, the design scheme does not satisfy the design requirements. Due to many factors, such as the earth's oblateness and the influence of the low-orbit atmosphere, there is a certain drift in the orbital height. On account of the effect of machining accuracy, there is also a slight random deviation in the focal length of the CCD camera. Therefore, the parameters $h_s$ and $f_c$ are modeled to be the normal random variables [32]. Besides, the parameters $b_{sat}$, $l_{sat}$, and $l_{sat}$ are the normal random variables [32] by reason of the machining errors in the structure. In addition to the above five parameters, the data handling (DH) subsystem quality estimation coefficient $c_{DH}$, and the telemetry, tracking and command subsystem (TTC) quality estimation coefficient $c_{TTC}$ also affect the microsatellite mass. Because the designers have insufficient information about the DH and TTC subsystems, the values of the coefficients $c_{DH}$ and $c_{TTC}$ cannot be obtained exactly. The above seven parameters' statistical properties are shown in Table 4.

**Table 4** Seven random variables that effect the conceptual microsatellite mass.

| Random variable | Mean (Lower boundary) | Standard deviation (Upper boundary) | Distribution |
| --- | --- | --- | --- |
| $h_o$ | 600 km | 6 km | Normal |
| $f_c$ | 280 mm | 1 mm | Normal |
| $b_{sat}$ | 800 mm | 10 mm | Normal |
| $l_{sat}$ | 700 mm | 10 mm | Normal |
| $t_{sat}$ | 5 mm | 0.1 mm | Normal |
| $c_{DH}$ | 0.04 | 0.05 | Uniform |
| $c_{TTC}$ | 0.05 | 0.06 | Uniform |

*5.1.2. Constructing Deep PCNN and preparing data for application 1*

In the *application 1*, the adaptive expansion coefficients of the auxiliary model are solved by a DNN with 36 outputs and five hidden layers, where the neuron numbers of five hidden layers are 64, 128, 128, 128, and 128, respectively. The DNN adopts the activation function $\text{ReLU}(x)$. Besides, the main model is a 5-order, 6-order, 7-order, or 8-order polynomial chaos neural network. For four polynomial chaos neural networks, the numbers of neurons in the orthogonal polynomial neural layer are 792, 1716, 3432, and 6435, respectively. The dataset $\mathcal{D}_{ce}$ consists of $2\times10^5$ unlabeled data, and the number $N_{gd}$ of labeled data is 50, 70, 90, and 150, respectively. The maximum training epoch $ep_{\max}$ is set to be 20000.

*5.1.3. Result analysis and discussion of application 1*

By the trained Deep PCNN, the first four statistical moments and their relative errors are shown in Table 5, and the bold results indicate the best results. Besides, this engineering problem is also solved by the OLS-PCE, OMP-SPCE, and LAR-SPCE. The results by the MCS method ($5\times10^5$ runs) are assumed to be the real results.

**Table 5** The first four statistical moments and their relative errors in the *Application 1*.

| $p$ | Method | $N_{gd}$ | Mean | S.D. | Skewness (R.E.) | Kurtosis (R.E.) |
|---|---|---|---|---|---|---|
| - | MCS | $5\times10^5$ | 176.24 | 2.52 | 0.0464 | 3.0034 |
| 5 | OLS-PCE | 1000 | 176.24 | 2.52 | 0.0498 (7.1637%) | 3.0092 (0.1933%) |
|   | OMP-SPCE | 500 | 176.24 | 2.52 | 0.0435 (6.6356%) | 3.0052 (0.0588%) |
|   | LAR-SPCE | 200 | 176.24 | 2.52 | 0.0385 (17.1328%) | 3.0063 (0.0968%) |
|   | **Deep PCNN** | **50** | 176.24 | 2.52 | **0.0458 (1.3374%)** | **3.0025 (0.0297%)** |
| 6 | OLS-PCE | 3500 | 176.24 | 2.52 | 0.0503 (8.0721%) | 3.0076 (0.1387%) |
|   | OMP-SPCE | 1200 | 176.24 | 2.52 | 0.0387 (16.7120%) | 2.9996 (0.1268%) |
|   | LAR-SPCE | 340 | 176.24 | 2.52 | 0.0407 (12.3542%) | 3.0020 (0.0473%) |
|   | **Deep PCNN** | **70** | 176.24 | 2.52 | **0.0499 (7.380%)** | **3.0029 (0.0173%)** |
| 7 | OLS-PCE | 6000 | 176.24 | 2.52 | 0.0423 (8.880%) | 3.0229 (0.6507%) |
|   | OMP-SPCE | 2800 | 176.24 | 2.52 | 0.0391 (15.7727%) | 3.0010 (0.0804%) |
|   | LAR-SPCE | 1100 | 176.24 | 2.52 | 0.0388 (16.3974%) | 3.0014 (0.0665%) |
|   | **Deep PCNN** | **90** | 176.24 | 2.52 | **0.0497 (7.1240%)** | **3.0031 (0.0093%)** |
| 8 | OLS-PCE | 15000 | 176.24 | 2.52 | 0.0448 (3.6559%) | 3.0024 (0.0317%) |
|   | OMP-SPCE | 6000 | 176.24 | 2.52 | 0.0425 (8.4810%) | 3.0013 (0.0681%) |
|   | LAR-SPCE | 2000 | 176.24 | 2.52 | 0.0447 (3.6974%) | 3.0024 (0.0343%) |
|   | **Deep PCNN** | **150** | 176.24 | 2.52 | **0.0450 (3.0250%)** | **3.0026 (0.0269%)** |

For $p=5$, although four methods can correctly calculate the mean and the standard deviation of the performance function value, the relative errors of three existing PCE methods on the skewness and the kurtosis are larger than the proposed Deep PCNN. Among three existing PCE methods, the OMP-SPCE ($N_{gd}=500$) has the minimal relative errors on the skewness (6.6356%) and the kurtosis (0.0588%), while the relative error on the skewness and the kurtosis of the proposed Deep PCNN ($N_{gd} = 90$) are only 1.3374% and 0.0297%, respectively. Besides, the LAR-SPCE uses the minimum labeled data in three existing PCE methods, i.e., $N_{gd} = 200$. However, the proposed Deep PCNN only uses 90 labeled data. What's more, the relative errors on the skewness (17.1328%) and the kurtosis (0.0968%) of the LAR-SPCE are far larger than the proposed Deep PCNN. For $p = 6, 7, 8$, the relative errors on the skewness and the kurtosis of the proposed Deep PCNN are always minimum in all methods. It is noteworthy that the proposed Deep PCNN always uses the fewest labeled data.

To further validate the accuracy of the proposed Deep PCNN, four error indicators, the failure probability $P_f$ and its relative estimation error rate $\varepsilon_{P_f}$ are also calculated as shown in Table 6. For $p = 5$, the minimum *RMSE*, *MAE*, and *MRE* in three existing PCE methods are 0.0227, 0.0169, and $0.9567 \times 10^{-3}$, respectively. However, the *RMSE*, *MAE*, and *MRE* of the proposed Deep PCNN are only 0.0101, 0.0079, and $0.4511 \times 10^{-3}$, respectively. Apparently, the *RMSE*, *MAE*, and *MRE* of the proposed Deep PCNN are smaller than three existing PCE methods. The maximum $R^2$ of three existing PCE methods is 0.999919, while the proposed Deep PCNN is 0.999984. Compared with the failure probability estimated by MCS, the relative estimation error rates of the failure probability $P_f$ for four methods are 3.1579% (LAR-SPCE), 0.5742% (OLS-PCE), 0.3349% (Deep PCNN), 0.0957% (OMP-SPCE). Although the $\varepsilon_{P_f}$ of the proposed Deep PCNN is slightly more than the OMP-SPCE, the OMP-SPCE uses ten times as much labeled data as the proposed Deep PCNN. Besides, the OMP-SPCE needs the complex orthogonal basis sparse operation while the proposed Deep PCNN does not. Thus, the proposed Deep PCNN can use fewer labeled data to obtain the failure probability $P_f$ close to the results of MCS.

For $p = 6, 7$, the *RMSE*, *MAE*, and *MRE* of the proposed Deep PCNN are still minimum in all methods, and the proposed Deep PCNN still has the largest $R^2$. For $p = 8$, the *RMSE*, *MAE*, *MRE*, $R^2$, and $\varepsilon_{P_f}$ of the proposed Deep PCNN ($N_{gd} = 150$) and the LAR-SPCE ($N_{gd} = 2000$) are very close. However, the LAR-SPCE adopts more than thirteen times as much labeled data as the proposed Deep PCNN. Based on the kernel density estimation (KDE), the estimated probability density functions (PDFs) of the satellite mass by the proposed *p*-order ($p = 5, 6, 7, 8$) Deep PCNN and the MCS are shown in Fig. 9. The red curves (solid line) are the results of the proposed Deep PCNN, and the green curves (dash line) are the results of the MCS. Referring to Fig. 9, the PDF curves of the satellite mass by the proposed *p*-order ($p = 5, 6, 7, 8$) Deep PCNN are very closer to the results of the MCS.

In summary, the proposed Deep PCNN can use less labeled data to build a more accurate surrogate model without any complex orthogonal basis sparse operation than the existing PCE methods for the *Application 1*. According to the estimated failure probability in Table 6, the conceptual design scheme of the studied microsatellite has high reliability.

**Table 6** Four error indicators and the estimated failure probability in the *Application 1*.

| $p$ | Method | $N_{gd}$ | RMSE | MAE | MRE ($\times 10^{-4}$) | $R^2$ | $P_f$ ($\times 10^{-3}$) | $\varepsilon_{P_f}$ |
|---|---|---|---|---|---|---|---|---|
| - | MCS | $5\times 10^5$ | - | - | - | - | 4.1800 | - |
| 5 | OLS-PCE | 1000 | 0.0477 | 0.0228 | 1.2929 | 0.999641 | 4.2040 | 0.5742% |
|   | OMP-SPCE | 500 | 0.0245 | 0.0180 | 1.0225 | 0.999905 | **4.1840** | **0.0957%** |
|   | LAR-SPCE | 200 | 0.0227 | 0.0169 | 0.9567 | 0.999919 | 4.0480 | 3.1579% |
|   | **Deep PCNN** | **50** | **0.0101** | **0.0079** | **0.4511** | **0.999984** | 4.1660 | 0.3349% |
| 6 | OLS-PCE | 3500 | 0.0334 | 0.0142 | 0.8072 | 0.999824 | **4.2480** | **1.6268%** |
|   | OMP-SPCE | 1200 | 0.0232 | 0.0180 | 1.0196 | 0.999915 | 4.0960 | 2.0096% |
|   | LAR-SPCE | 340 | 0.0208 | 0.0153 | 0.8668 | 0.999932 | 4.0880 | 2.2010% |
|   | **Deep PCNN** | **70** | **0.0156** | **0.0117** | **0.6646** | **0.999961** | 4.2580 | 1.8660% |
| 7 | OLS-PCE | 6000 | 0.0710 | 0.0212 | 1.2055 | 0.999206 | 4.2180 | 0.9091% |
|   | OMP-SPCE | 2800 | 0.0318 | 0.0248 | 1.4053 | 0.999841 | 4.1000 | 1.9139% |
|   | LAR-SPCE | 1100 | 0.0193 | 0.0148 | 0.8392 | 0.999941 | 4.0860 | 2.2488% |
|   | **Deep PCNN** | **90** | **0.0121** | **0.0093** | **0.5267** | **0.999977** | 4.1960 | 0.3828% |
| 8 | OLS-PCE | 15000 | 0.0150 | 0.0674 | 0.8535 | 0.999286 | 4.1280 | 1.2440% |
|   | OMP-SPCE | 6000 | 0.0303 | 0.0236 | 1.3379 | 0.999855 | 4.1100 | 1.6746% |
|   | LAR-SPCE | 2000 | **0.0109** | 0.0086 | 0.4907 | **0.999981** | 4.1660 | 0.3349% |
|   | **Deep PCNN** | **150** | 0.0114 | **0.0083** | **0.4720** | 0.999979 | 4.1640 | 0.3828% |

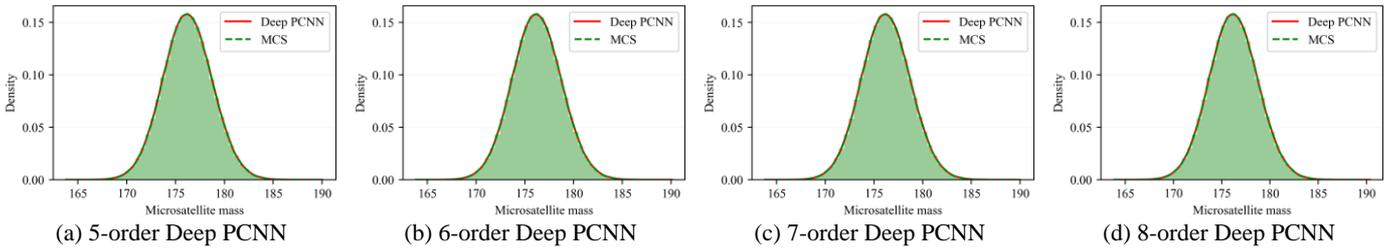

(a) 5-order Deep PCNN  (b) 6-order Deep PCNN  (c) 7-order Deep PCNN  (d) 8-order Deep PCNN

**Fig. 9** The estimated probability density functions of the satellite mass by the proposed *p*-order ($p = 5, 6, 7, 8$) Deep PCNN and the MCS.

*5.2. Application 2: First-order frequency reliability analysis of "Tiantuo-3" satellite load-bearing structure*

*5.2.1. Engineering background of application 2*

As shown in Fig. 10 (a), "Tiantuo-3" satellite is a highly integrated microsatellite with low mass. For its load-bearing structure, including frame structure, support rod, and separating device in Fig. 10 (b), the strength design is important for developing the "Tiantuo-3" satellite. If the "Tiantuo-3" satellite and the launch vehicle resonate during the launch process, the former's load-bearing structure will be broken, causing the satellite to fail. To avoid resonance, the first-order frequency of the load-bearing structure is usually more than 81.7925 Hz.

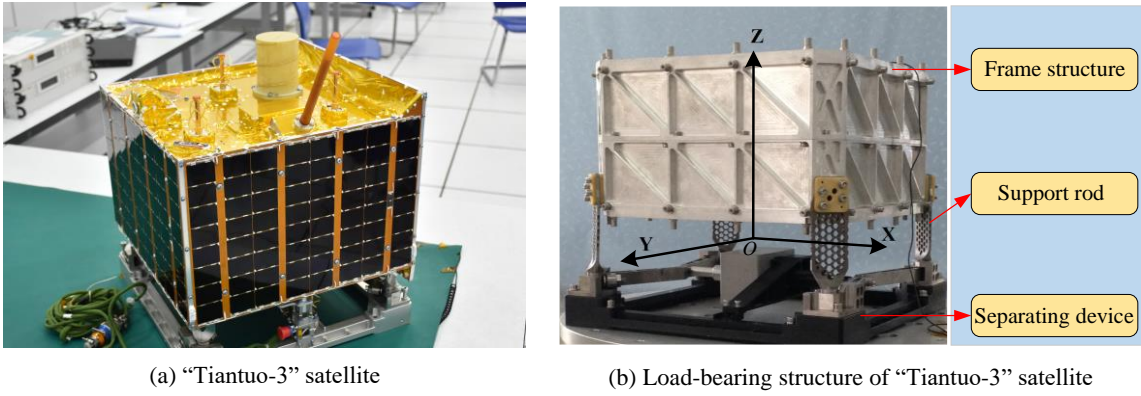

(a) "Tiantuo-3" satellite    (b) Load-bearing structure of "Tiantuo-3" satellite

**Fig. 10** "Tiantuo-3" satellite and its load-bearing structure.

For the "Tiantuo-3" satellite's load-bearing structure, it is mainly determined by six physical parameters, i.e., the aluminum alloy density $\rho_{AE}$, the spring steel density $\rho_{SS}$, the titanium alloy density $\rho_{TA}$, the aluminum elastic modulus $E_{AE}$, the spring steel elastic modulus $E_{SS}$, and the titanium alloy elastic modulus $E_{TA}$. Due to uncertainty factors existing in the ambient temperature, material molding conditions, etc., the above six physical parameters are modeled as six random variables [21], as shown in Table 7.

**Table 7** Six random variables involved in the "Tiantuo-3" satellite's load-bearing structure

| Random variable | Mean | Standard deviation | Distribution |
| --- | --- | --- | --- |
| $\rho_{AE}$ | $2.69\,\text{g}/\text{cm}^3$ | $8.97\times10^{-3}\,\text{g}/\text{cm}^3$ | Normal |
| $\rho_{SS}$ | $7.85\,\text{g}/\text{cm}^3$ | $2.617\times10^{-2}\,\text{g}/\text{cm}^3$ | Normal |
| $\rho_{TA}$ | $4.43\,\text{g}/\text{cm}^3$ | $1.477\times10^{-2}\,\text{g}/\text{cm}^3$ | Normal |
| $E_{AE}$ | $6.89\times10^{4}\,\text{MPa}$ | $2.29667\times10^{3}\,\text{MPa}$ | Normal |
| $E_{SS}$ | $2.00\times10^{5}\,\text{MPa}$ | $2.00\times10^{3}\,\text{MPa}$ | Normal |
| $E_{TA}$ | $1.138\times10^{5}\,\text{MPa}$ | $1.89667\times10^{3}\,\text{MPa}$ | Normal |

*5.2.2. Constructing Deep PCNN and preparing data for application 2*

In the *application 2*, the adaptive expansion coefficients of the auxiliary model are solved by a DNN with 28 outputs and five hidden layers, where the neuron numbers of five hidden layers are 32, 64, 128, 64, and 64, respectively. The DNN adopts the activation function $\text{ReLU}(x)$. Besides, the main model is a 6-order, or 8-order polynomial chaos neural network. For two polynomial chaos neural networks, the numbers of neurons in the orthogonal polynomial neural layer are 924 and 3003, respectively. The dataset $\mathcal{D}_{ce}$ consists of $4\times10^5$ unlabeled data, and the number $N_{gd}$ of labeled data is 60 and 70, respectively. The maximum training epoch $ep_{\max}$ is set to be 20000.

*5.2.3. Result analysis and discussion of Application 2*

The first four statistical moments and their relative errors are calculated by the trained Deep PCNN, as shown in Table 8. Besides, this engineering problem is also solved by the SP-SPCE, OMP-SPCE, and LAR-SPCE. Due to the limited experimental data collected, this paper takes the results of 600 MCSs as the real results. In Table 8, the bold results indicate the best results.

For $p=6$, although four methods can correctly calculate the mean and the standard deviation of the first-order frequency, the relative errors of three existing PCE methods on the skewness and the kurtosis are larger than the proposed Deep PCNN. Among three existing PCE methods, the minimal relative errors on the skewness and the kurtosis are 1.0148% and 0.0169%, respectively. In contrast, the relative error on the skewness and the kurtosis of the proposed Deep PCNN ($N_{gd}=60$) are only 0.8673% and 0.0042%, respectively. Besides, the SP-SPCE uses the fewest labeled data in three existing PCE methods, i.e., $N_{gd}=100$. However, the proposed Deep PCNN only uses 60 labeled data. What's more, the relative errors on the skewness (1.0798%) and the kurtosis (0.0581%) of the SP-SPCE are larger than the proposed Deep PCNN. For $p=8$, the relative errors on the skewness and the kurtosis of the proposed Deep PCNN are also minimum in all methods. It is noteworthy that the proposed Deep PCNN still uses the fewest labeled data.

**Table 8** The first four statistical moments and their relative errors in the *Application 2*.

| $p$ | Method | $N_{gd}$ | Mean | Standard deviation | Skewness (R.E.) | Kurtosis (R.E.) |
|---|---|---|---|---|---|---|
| - | MCS | 600 | 82.836 | 0.453 | -0.07256 | 2.6518 |
| 6 | LAR-SPCE | 140 | 82.836 | 0.453 | -0.07138 (1.6086%) | 2.6513 (0.0169%) |
|   | OMP-SPCE | 130 | 82.836 | 0.453 | -0.073282 (1.0148%) | 2.6522 (0.0184%) |
|   | SP-SPCE | 100 | 82.836 | 0.453 | -0.073329 (1.0798%) | 2.6533 (0.0581%) |
|   | **Deep PCNN** | **60** | 82.836 | 0.453 | **-0.07319 (0.8673%)** | **2.6516 (0.0042%)** |
| 8 | LAR-SPCE | 180 | 82.836 | 0.453 | -0.07219 (0.4908%) | 2.6515 (0.0929%) |
|   | OMP-SPCE | 170 | 82.836 | 0.453 | -0.07216 (0.5366%) | 2.6523 (0.0309%) |
|   | SP-SPCE | 140 | 82.836 | 0.453 | -0.07162 (1.2782%) | 2.6524 (0.0231%) |
|   | **Deep PCNN** | **70** | 82.836 | 0.453 | **-0.07229 (0.3773%)** | **2.6517 (0.0030%)** |

To further validate the accuracy of the proposed Deep PCNN, four error indicators, the failure probability $P_f$ and its relative estimation error rate $\varepsilon_{P_f}$ are also calculated as shown in Table 9, and , the bold results indicate the best results. For $p=6$, the minimum *RMSE*, *MAE*, and *MRE* in three existing PCE methods are $6.0706\times10^{-4}$, $4.0844\times10^{-4}$, and $0.0493\times10^{-4}$, respectively. However, the *RMSE*, *MAE*, and *MRE* of the proposed Deep PCNN are only $5.4234\times10^{-4}$, $3.7483\times10^{-4}$, and $0.0452\times10^{-4}$, respectively. Thus, the *RMSE*, *MAE*, and *MRE* of the proposed Deep PCNN are smaller than three existing PCE methods. The maximum $R^2=0.9999982$ of three existing PCE methods is still less than the $R^2=0.9999986$ of proposed Deep PCNN. Compared with the failure probability estimated by MCS, three existing PCE methods have a 20% relative estimation error rate, while the proposed Deep PCNN can estimate the failure probability accurately. The labeled data used by the proposed Deep PCNN is also less than three existing PCE methods. Besides, three existing PCE methods need the complex orthogonal basis sparse operation while the proposed Deep PCNN does not. For $p=8$, the *RMSE*, *MAE*, and *MRE* of the proposed Deep PCNN are always minimum in all methods, and the proposed Deep PCNN still has the largest $R^2$. All methods can estimate the failure probability accurately.

**Table 9** Four error indicators and the estimated failure probability in the *Application 2*.

| $p$ | Method | $N_{gd}$ | RMSE ($\times 10^{-4}$) | MAE ($\times 10^{-4}$) | MRE ($\times 10^{-4}$) | $R^2$ | $P_f$ | $\varepsilon_{P_f}$ |
|---|---|---|---|---|---|---|---|---|
| - | MCS | 600 | - | - | - | - | 0.0083 | - |
| 6 | LAR-SPCE | 140 | 7.3182 | 4.0844 | 0.0493 | 0.9999974 | 0.0067 | 20.00% |
|   | OMP-SPCE | 130 | 6.0706 | 4.3101 | 0.0521 | 0.9999982 | 0.0067 | 20.00% |
|   | SP-SPCE | 100 | 8.1117 | 4.5800 | 0.0554 | 0.9999968 | 0.0067 | 20.00% |
|   | **Deep PCNN** | **60** | **5.4234** | **3.7483** | **0.0452** | **0.9999986** | **0.0083** | **0.00%** |
| 8 | LAR-SPCE | 180 | 8.3160 | 4.2878 | 0.0518 | 0.9999966 | 0.0083 | 0.00% |
|   | OMP-SPCE | 170 | 6.7627 | 4.4724 | 0.0540 | 0.9999978 | 0.0083 | 0.00% |
|   | SP-SPCE | 140 | 7.8291 | 4.4326 | 0.0535 | 0.9999970 | 0.0083 | 0.00% |
|   | **Deep PCNN** | **70** | **5.9560** | **4.1652** | **0.0503** | **0.9999983** | **0.0083** | **0.00%** |

In summary, the result analyses for Table 8 and Table 9 show that the proposed Deep PCNN can use less labeled data to build a more accurate surrogate model without any complex orthogonal basis sparse operation than the existing PCE methods. To more accurately estimate the failure probability, $1\times 10^6$ MCSs are performed on the trained 8-order Deep PCNN. The estimated failure probability is 0.008405. Therefore, the "Tiantuo-3" satellite load-bearing structure has high first-order frequency reliability.

## 6. Conclusions

This paper proposes a consistency regularization-based deep polynomial chaos neural network (Deep PCNN) method for analyzing stochastic system reliability. The Deep PCNN consists of two models, i.e., the auxiliary and the main models. The main model is a polynomial chaos neural network built by a high-order PCE model. In particular, the expansion coefficients of the high-order PCE model are parameterized into the learnable weights of the polynomial chaos neural network, realizing iterative learning of expansion coefficients to obtain more accurate high-order PCE models. The auxiliary model is constructed to be a low-order adaptive PCE model by a DNN, where the DNN solves the expansion coefficients of the low-order adaptive PCE model. The auxiliary model uses the proposed consistency regularization loss function to assist in training the main model. Based on a small amount of labeled data and abundant unlabeled data, the consistency learning algorithm is proposed to learn the parameters of the Deep PCNN. After training the Deep PCNN, the MCS can be straightforwardly performed on the main model to analyze the system reliability. This paper adopts a numerical example to validate the effectiveness of the consistency regularization-based Deep PCNN method, and the results show that the consistency regularization-based Deep PCNN method can significantly reduce the number of labeled data in constructing a high-order PCE model without losing accuracy. Finally, the consistency regularization-based Deep PCNN method is applied to analyze the reliability of two aerospace engineering systems. Next, the authors will study how to reduce the amount of labeled data in building the PCE models of the high dimensional stochastic systems in future research.

**Acknowledgments**

This work was supported by the Postgraduate Scientific Research Innovation Project of Hunan Province (No.CX20200006) and the National Natural Science Foundation of China (Nos.11725211).